\documentclass[10pt,twocolumn,letterpaper]{article}

\usepackage[pagenumbers]{cvpr} 

\usepackage{amsthm}
\usepackage{float}
\floatstyle{ruled}\newfloat{algorithm}{tbp}{loa}\floatname{algorithm}{Algorithm}
\newcommand{\kw}[1]{\textbf{#1}}
\newcommand{\cmt}[1]{\hfill{\footnotesize\color{gray}$\triangleright$ #1}}
\usepackage{xspace}
\usepackage{tikz}
\usetikzlibrary{positioning,arrows.meta,calc}
\usepackage{microtype}

\theoremstyle{plain}
\newtheorem{lemma}{Lemma}
\newtheorem{proposition}{Proposition}
\theoremstyle{remark}

\newcommand{\R}{\mathbb{R}}
\newcommand{\E}{\mathbb{E}}
\newcommand{\Prob}{\mathbb{P}}
\newcommand{\ip}[2]{\langle #1, #2\rangle}
\newcommand{\norm}[1]{\lVert #1\rVert}
\newcommand{\grs}{GRS\xspace}
\newcommand{\grsz}{GRS$_0$\xspace}
\newcommand{\saa}{sAA\xspace}
\newcommand{\spgd}{sPGD\xspace}
\newcommand{\srs}{Sparse-RS\xspace}
\newcommand{\Tadd}{T_{\mathrm{add}}}
\newcommand{\Tdrop}{T_{\mathrm{drop}}}

\makeatletter
\@ifpackagelater{lineno}{2026/02/01}{%
  \RemoveFromHook{build/column/before}[lineno]%
  \AtBeginDocument{%
    \def\@LN@makecol{%
      \@LN@orig@makecol
      \setbox\@outputbox \vbox{%
        \boxmaxdepth \@maxdepth
        \protected@write\@auxout{}{\string\def\string\@LN@column{\if@firstcolumn1\else2\fi}}%
        \box\@outputbox
      }%
    }%
    \let\@makecol\@LN@makecol
  }%
}{}
\makeatother

\definecolor{cvprblue}{rgb}{0.21,0.49,0.74}
\usepackage[pagebackref,breaklinks,colorlinks,allcolors=cvprblue]{hyperref}

\def\confName{CVPR}
\def\confYear{2027}

\title{Propose with the Gradient, Accept with the Loss:\\ One Hedged Attack for Pixel-Sparse Robustness}

\author{Hon Ming Yam$^{1}$ \quad Chun Pong Lau$^{1}$\\
$^{1}$City University of Hong Kong \quad 
{\tt\small \{hmyam4, cplau27\}@cityu.edu.hk}
}

\begin{document}
\maketitle
\begin{abstract}
A pixel-sparse adversary may change a few pixels of an image to any value.
Robustness to such attacks is currently measured with an ensemble, sparse-AutoAttack (\saa), because its members fail in different places: sparse projected gradient descent (\spgd) trusts a first-order model that a full pixel change breaks, and sparse random search (\srs) ignores the gradient even where it is informative.
We show that both attacks are local searches, each a proposal distribution plus an acceptance rule, and ask what such a search can learn from the gradient.
A Taylor bound says what to measure, and measurements at the states the search visits give five rules.
The sign of the gradient is its most reliable use at a single pixel and sets a new pixel's value.
Its ranking of candidate pixels should be followed loosely, and its ranking of the pixels to remove, which is often more reliable, can be followed more closely.
A pixel should not be proposed twice before a move is accepted, and acceptance should be decided by the loss alone.
These rules define Gradient-guided Random Search (\grs), which runs once and costs one forward and one backward pass per step.
On 13 CIFAR-10 configurations and 5 ImageNet models, \grs matches or beats the best single member of \saa and is within one point of \saa itself or stronger on 17 of 18 settings, and it is significantly stronger than \saa on 8.
It uses 3 instead of up to 7 forward-pass units, and on CIFAR-10 it runs 3.7 to 5.0 times faster than the members of \saa run in full.
\end{abstract}

\section{Introduction}
\label{sec:intro}

\begin{figure}[t]
\centering
\includegraphics[width=\linewidth]{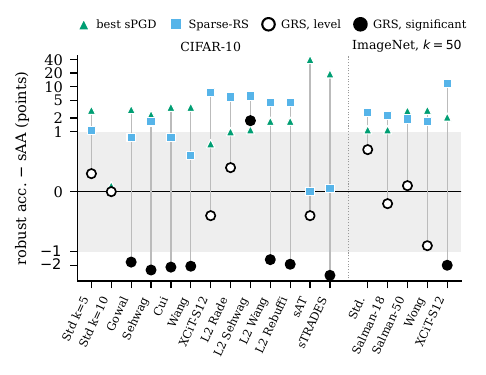}
\caption{Robust accuracy minus that of \saa on the 18 main settings (symmetric log scale; below 0 is stronger than \saa; grey band $\pm 1$ point). Filled: \grs differs significantly from \saa (exact McNemar, $p < 0.05$). \grs costs 3 units per point, \saa up to 7.}
\label{fig:teaser}
\end{figure}

A pixel-sparse adversary may change at most $k$ pixels of an image, and each changed pixel may take any value in the image domain.
This threat model is hard to evaluate: the constraint set is non-convex, a single admissible pixel change is not small, and the first-order tools of $\ell_\infty$ and $\ell_2$ attacks~\cite{goodfellow2015explaining,madry2018towards} have no direct counterpart.
For these reasons the current evaluation standard, sparse-AutoAttack (\saa)~\cite{zhong2024sparse}, is an ensemble.
It runs two white-box attacks, sparse projected gradient descent (\spgd) with a projected and with an unprojected gradient, and one black-box attack, sparse random search (\srs)~\cite{croce2022sparsers}, and it counts a point as broken if any member breaks it.
At 10k iterations the ensemble costs up to 7 units of 10k forward passes per point: 3 for each \spgd run (a backward pass counts as two) and 1 for \srs, with later members of the cascade run only on points still unbroken.
Its members are not redundant: on the models we study, each of the three is the best one somewhere, and each is far behind the ensemble somewhere (\cref{fig:teaser,tab:main-cifar}).

This paper asks why the ensemble is needed and whether a single attack can replace it.
Our starting point is that every pixel-$\ell_0$ attack, white-box or black-box, is a \emph{local search} over the pair (which pixels, which values).
A proposal distribution $Q$ suggests a candidate from the current state, and an acceptance rule decides whether to move to it.
Seen this way, the two kinds of member of \saa misuse the gradient in opposite ways.
\spgd accepts every iterate, so it has to trust first-order \emph{magnitudes}, and it collapses where they mislead.
On the models trained with sparse adversarial training (sAT) and its TRADES~\cite{zhang2019theoretically} variant (sTRADES)~\cite{zhong2024sparse} it leaves 75--84\% of the points unbroken, against 35--61\% for \srs (\cref{tab:main-cifar}).
\srs chooses pixels blindly and fills them with random values, so it is slow where the gradient is informative.
On an ImageNet XCiT it leaves 40.3\% unbroken at $k = 50$, against 30.5\% for \spgd (\cref{tab:main-imagenet}).
Neither attack sets a newly added pixel directly to the corner of its box given by the sign of the gradient, which is the most reliable first-order quantity at a single pixel.
\saa hedges between the two behaviours by running all three members.

We make the hedge explicit and put it inside one run.
A Taylor bound (\cref{lem:reliability}) shows that the first-order prediction for a full pixel change carries an error of the size of the local curvature.
Whether the gradient can be trusted is therefore a ratio of predicted gain to curvature, which can be measured rather than assumed.
We measure it at the states a search actually visits, on 12 CIFAR-10 and 5 ImageNet models, and the measurements give five design rules (\cref{sec:rules}).
Accept a move only if the true loss improves.
Choose pixels to add from a \emph{tempered} gradient ranking, whose top is unreliable on robust ImageNet models, and pixels to drop from a \emph{colder} ranking, because a pixel's current contribution is often far more reliable than a candidate's predicted gain.
Set a new pixel's value from the gradient's sign, and spend no further steps on values.
Do not propose a pixel twice before the next move is accepted.
The attack that follows these rules, Gradient-guided Random Search (\grs, \cref{sec:method}), is a random search like \srs whose moves are drawn from the gradient and accepted by the loss.
It costs one forward and one backward pass per step, the same as \spgd.

Our contributions are:
\begin{itemize}[leftmargin=*,itemsep=1pt,topsep=2pt]
\item a local-search view of pixel-$\ell_0$ attacks, with elementary bounds that separate the two ways of failing and bound the cost of a tempered proposal (\cref{sec:search});
\item a measurement of what the gradient can be trusted for at the states a search visits, which turns into five design rules and one attack (\cref{sec:rules,sec:method});
\item an evaluation on 13 CIFAR-10 configurations (12 models) and 5 ImageNet models at three budgets, against the \saa members, $\sigma$-zero~\cite{cina2025sigmazero}, PGD$_0$~\cite{croce2019sparse} and GreedyPixel~\cite{wang2025greedypixel} (\cref{sec:experiments}): on 17 of 18 main settings \grs matches or beats the best single member and is within one point of \saa or stronger, and it is significantly stronger than \saa on 8, at 3 instead of up to 7 units;
\item an ablation of the design choices, in which greedy acceptance has by far the largest effect, and diagnostics that explain most of the differences between models (\cref{sec:ablation}).
\end{itemize}

\section{Related work}
\label{sec:related}

\paragraph{Sparse adversarial attacks.}
Adversarial examples~\cite{szegedy2014intriguing} with few changed pixels were first found by greedy saliency methods (the Jacobian-based saliency map attack, JSMA~\cite{papernot2016limitations}, GreedyFool~\cite{dong2020greedyfool}), by evolutionary search (the one-pixel attack~\cite{su2019onepixel}) or by relaxations of the constraint: Carlini and Wagner~\cite{carlini2017towards} prune an $\ell_2$ solution, SparseFool~\cite{modas2019sparsefool} linearizes the decision boundary, PGD$_0$~\cite{croce2019sparse} projects onto the $\ell_0$ ball, $\sigma$-zero~\cite{cina2025sigmazero} minimizes a smooth surrogate of the $\ell_0$ norm, boundary attacks~\cite{brendel2019accurate,pintor2021fast} minimize it directly, and EAD~\cite{chen2018ead} and $\ell_1$-APGD~\cite{croce2021mind} attack the $\ell_1$ relaxation.
The relaxations take a dense gradient step and then make it sparse, so they trust the first-order model for exactly the full pixel changes where \cref{lem:reliability} says it can fail.
SparseFool and $\sigma$-zero also minimize the size of the perturbation rather than succeed within a fixed budget, and at the budgets used for evaluation $\sigma$-zero and PGD$_0$ leave 1.8 to 5.8 times as many points unbroken as \saa on the RobustBench CIFAR-10 models (\cref{tab:main-cifar}).

Sparse-PGD~\cite{zhong2024sparse,zhong2025sparsepgd} is the strongest white-box attack for a fixed pixel budget.
It relaxes the set of changed pixels to continuous mask logits, alternates projected gradient steps on the mask and on the pixel values, and redraws the mask when progress stalls.
It accepts every iterate, so it has no defence against a misleading gradient: on sAT and sTRADES, which are trained against \spgd itself, it leaves far more points unbroken than \srs (\cref{tab:main-cifar}).
Neither of its projected and unprojected variants is the stronger one on every model, which is why \saa runs both.
Among black-box attacks, \srs~\cite{croce2022sparsers} extends the Square Attack~\cite{andriushchenko2020square} to a random search over pixel sets with greedy acceptance and a shrinking number of swapped pixels, and also covers patches~\cite{brown2017adversarial,karmon2018lavan}; Pixle~\cite{pomponi2022pixle} instead rearranges pixels.
Greedy acceptance makes \srs robust to gradient masking, but it chooses pixels uniformly and gives each a random value, so it needs many queries where the gradient would point the way, and on ImageNet at $k = 50$ it is behind the projected \spgd on three of five models, by 9.8 points on XCiT (\cref{tab:main-imagenet}).
GreedyPixel~\cite{wang2025greedypixel} sets pixels one at a time in an order fixed by gradient saliency at the clean input and never revises that order.
\saa~\cite{zhong2024sparse} combines two \spgd variants with \srs in the spirit of AutoAttack~\cite{croce2020reliable}; sAT and sTRADES apply adversarial training~\cite{madry2018towards} with \spgd.
The ensemble covers each member's failures at up to 7 units per point, but no member combines the gradient's guidance with greedy acceptance.

\paragraph{Local search and tempered proposals.}
Greedy acceptance of random swaps is hill climbing~\cite{aarts1997local,hoos2004stochastic}.
Proposing moves in proportion to the exponential of a score uses the Boltzmann weights of Metropolis sampling and simulated annealing~\cite{metropolis1953equation,kirkpatrick1983optimization}, and forbidding recently tried moves is tabu search~\cite{glover1989tabu}.
Drawing $q$ distinct pixels in proportion to $\exp(z/T)$ is done with the Gumbel-top-$q$ trick~\cite{kool2019stochastic}.

\paragraph{Gradient masking and ensembles.}
Adding a gradient-free attack to an ensemble is the standard remedy for gradient masking~\cite{uesato2018adversarial,athalye2018obfuscated,carlini2019evaluating,croce2020reliable,tramer2020adaptive}.
Certified $\ell_0$ defences~\cite{levine2020robustness,jia2022almost} bound robustness from below, attacks from above.
For $\ell_0$ attacks our analysis locates the failure in the acceptance rule rather than in the gradient itself.
With greedy acceptance on the true loss the search never moves to a worse state, and a misleading gradient lowers its acceptance rate by a bounded factor (\cref{prop:temper}), so one run can be both fast and safe.

\section{Pixel-$\ell_0$ attacks as local search}
\label{sec:search}

\paragraph{Setting.}
An input $x \in [0,1]^{C \times N}$ has $C$ channels, $N$ pixels and label $y$.
Write $f(\cdot)$ for the logits and $m(x') = f_y(x') - \max_{j \neq y} f_j(x')$ for the margin, so the attack succeeds exactly when $m(x') < 0$.
A \emph{state} $s = (S, v)$ consists of a \emph{support} $S$, the set of $k$ changed pixels, and a value $v_i$ for each $i \in S$.
The value lives in the box $B_i = \prod_c [-x_{ic}, 1 - x_{ic}]$, so that the perturbed pixel stays in $[0, 1]$, and the adversarial input is $x' = x + v$ on $S$.
Every attack in this paper repeats two steps: build a candidate $s' \sim Q(\cdot \mid s)$ from the current state, and apply an acceptance rule to decide whether $s'$ becomes the new state.
\spgd relaxes $S$ to continuous mask logits and takes projected gradient steps on mask and values, so its candidate is a deterministic function of the gradient, and it accepts every iterate.
\srs swaps $q$ uniformly chosen pixels, gives each new pixel a uniformly random corner of its box, and keeps the candidate only if its loss decreases (the margin, or the cross-entropy on robust models).

Let $g = \nabla m(x')$ be the gradient at the current input and $g_i \in \R^C$ its restriction to pixel $i$.
Two first-order scores will matter.
The \emph{add score} $\Phi_i = \max_{v \in B_i} \ip{-g_i}{v} \ge 0$ is the predicted decrease of the margin from adding a pixel $i \notin S$ at its best value.
That value is the \emph{best-response corner} $v_i^\ast$: in channel $c$ it equals $1 - x_{ic}$ if $g_{ic} < 0$ and $-x_{ic}$ otherwise.
The \emph{drop score} $\kappa_j = -\ip{g_j}{v_j}$ is the predicted contribution of a pixel $j \in S$, the first-order decrease of the margin it currently provides.

\paragraph{Single-pixel moves are not small.}
A pixel changes by $O(1)$ in every channel, so $\norm{v}^2 \le C$, and the first-order model has an error of the size of the local curvature rather than a vanishing one.
Let $\lambda_i$ be a Lipschitz constant of the gradient of $m$ with respect to pixel $i$'s $C$ coordinates on $B_i$.
Step-by-step proofs of all statements in this section and the next are in \cref{sec:proofs}.

\begin{lemma}[first-order reliability]
\label{lem:reliability}
Adding pixel $i$ at value $v \in B_i$, with the rest of the state fixed, changes the margin by $\Delta m$ with
$\lvert \Delta m - \ip{g_i}{v} \rvert \le \lambda_i C / 2$.
\end{lemma}

Two consequences follow.
\Cref{lem:reliability} certifies that adding pixel $i$ at $v_i^\ast$ lowers the margin if $\Phi_i > \lambda_i C/2$.
It certifies the predicted order of two pixels if their add scores differ by more than $(\lambda_i + \lambda_{i'}) C / 2$.
So whether the gradient can be trusted depends on the ratio $\Phi_i / \lambda_i$, and gradient masking is the extreme case $\Phi \approx 0$.
We do not estimate $\lambda_i$; \cref{sec:rules} measures reliability directly, as the correlation between first-order predictions and true margin changes.

\paragraph{A local model.}
To compare search strategies we use a simple additive model.
A set $\mathcal{H}$ of $\gamma$ \emph{helpful} pixels exists, the margin is $m(S) = m_0 - a\,\lvert S \cap \mathcal{H} \rvert$ when values are at their best, and the attack succeeds once $\lvert S \cap \mathcal{H} \rvert \ge r$, with $r \le k$ and $r \le \gamma$.
The search starts from a support with $j_0 < r$ helpful pixels.
Real margins are close to additive at the states we probe: the Pearson correlation between the sum of an add and a drop and the corresponding swap is 0.95--1.00 (\cref{tab:diag}).

\begin{proposition}[accepting every step]
\label{prop:accept}
(a) If $\Phi_i \le \lambda_i C / 2$ for every pixel, \cref{lem:reliability} certifies a decrease of $m$ for no single-pixel add, so an attack that accepts every iterate gets no descent certificate from it.
(b) If the gradients carry no information about $\mathcal{H}$, so that the sequence of supports is independent of a uniformly random $\mathcal{H}$, each evaluated support succeeds with probability at most $\binom{k}{r} (\gamma/N)^r$.
\end{proposition}

\begin{proposition}[greedy acceptance of blind proposals]
\label{prop:blind}
Under the additive model, greedy single-swap search with uniform proposals succeeds after an expected
$\E[\tau] = \sum_{j=j_0}^{r-1} \frac{k (N-k)}{(k-j)(\gamma-j)} \le \frac{r\,k\,(N-k)}{(k-r+1)(\gamma-r+1)}$
steps, whatever the gradient says.
\end{proposition}

For $N = 1024$, $k = 20$, $\gamma = 50$ and $r = 10$, \cref{prop:accept}(b) bounds a blind attack that accepts every step to one success in about $7 \times 10^{7}$ steps, whereas \cref{prop:blind} gives greedy uniform search about 300.
Greedy acceptance is what keeps a search alive when the gradient carries no information.

\section{What the gradient can be trusted for}
\label{sec:rules}

\begin{figure*}[t]
\centering
\includegraphics[width=\textwidth]{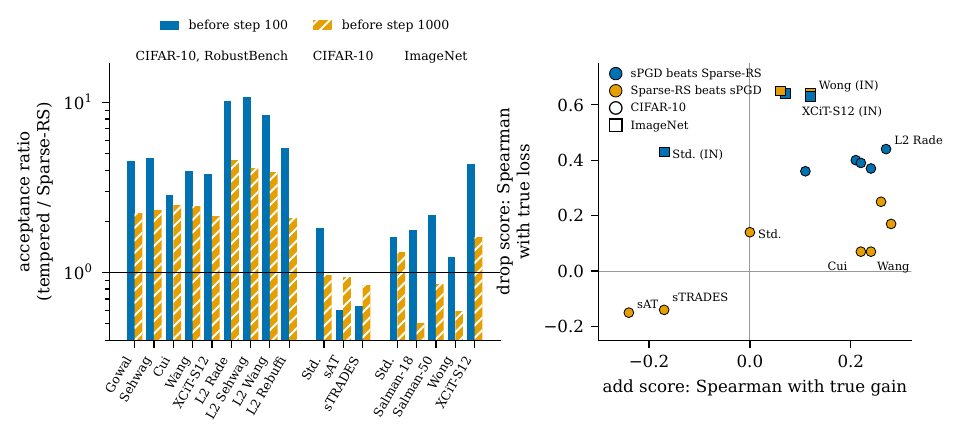}
\caption{Reliability at the states the search visits. \textbf{Left:} acceptance rate of a tempered gradient proposal (add and drop at $T=4$, best-response fill) divided by that of \srs's proposal (uniform swap, random corner), before steps 100 and 1000; log scale. Above 1 the gradient helps; on sAT, sTRADES and the robust ImageNet CNNs it does not before step 1000. \textbf{Right:} before step 1000, Spearman correlation of the add score $\Phi$ with the true gain of an add (x-axis) and of the drop score $\kappa$ with the true loss of a drop (y-axis), one marker per model. Colour shows which \saa member is stronger on the evaluation points: blue where the better \spgd run beats \srs, orange where \srs wins. (IN): ImageNet.}
\label{fig:diag}
\end{figure*}

The bounds of \cref{sec:search} turn reliability into something we can measure.
We measure it at the states a search actually visits.
We run a probe search on held-out points and stop it before steps 100 and 1000.
The probe proposes adds and drops at $T = 4$ and sets each added pixel at its best-response corner, as \grs does, but uses neither Rule 3 nor Rule 5 below.
The points are CIFAR-10 test points 1000--1999, on which no setting was tuned and nothing is evaluated, and ImageNet validation images 0--499, which are also used for the selection in \cref{sec:selection} but never for evaluation.
At each stopped state we evaluate, with forward passes, the true change of the margin for every single add at its best-response corner, for every single drop, and for samples of value changes and swaps, and we compare them with the first-order predictions.
\Cref{fig:diag} and \cref{tab:diag,tab:conc} (appendix) summarize the results for 12 CIFAR-10 and 5 ImageNet models.
Each finding below becomes a design rule.

\paragraph{Rule 1: accept with the loss, never with the gradient.}
First-order magnitudes are poor predictors everywhere.
Before step 1000, the Pearson correlation between the predicted and the true margin change of an add is 0.10--0.28 on the nine RobustBench CIFAR-10 models and $-0.21$ to $0.04$ on ImageNet.
On the standard CIFAR-10 model and on sAT and sTRADES it is negative ($-0.09$, $-0.29$ and $-0.13$), and on sAT and sTRADES even the rankings are reversed (\cref{fig:diag}, right, bottom left), so there the gradient is misleading rather than merely uninformative.
Acceptance must therefore be decided by the true loss, which costs one forward pass that the attack pays anyway, and never by a first-order prediction.
With greedy acceptance the search never moves to a worse state; this is the guarantee \spgd gives up.
Accepting every candidate instead leaves 9.1--29.2 points more unbroken (\cref{sec:ablation}).

\paragraph{Rule 2: choose pixels to add from a tempered ranking.}
Let $z(\cdot)$ standardize a score over the allowed pixels, so that it has mean 0 and variance 1 under the uniform distribution $U$, and let the proposal be $Q \propto \exp(z/T)$ with temperature $T$.

\begin{proposition}[tempered proposals]
\label{prop:temper}
Let $G$ be the set of moves that improve the current state and $w = Q/U = e^{z/T}/Z_T$ with $Z_T = \E_U[e^{z/T}]$. Under greedy acceptance, the acceptance probability satisfies $\Prob_Q(G) = \Prob_U(G)\, \E_U[w \mid G]$ and
\[
\exp\!\Big(\tfrac{\E_U[z \mid G]}{T} - \log Z_T\Big) \;\le\; \frac{\Prob_Q(G)}{\Prob_U(G)} \;\le\; \exp\!\Big(\tfrac{\max_G z}{T} - \log Z_T\Big).
\]
\end{proposition}

The proposition is stated for single moves drawn from one side; for a swap whose add and drop are drawn independently, $w$ is the product of the two sides' ratios and the exponents add.
Since $\E_U[z] = 0$, the normalizer satisfies $\log Z_T \ge 0$ (it equals $1/(2T^2)$ for Gaussian scores) and vanishes only as $T \to \infty$.
If the score is informative on average over the improving moves, that is $\mu_G = \E_U[z \mid G] > T \log Z_T$, the tempered proposal is accepted more often than a uniform one.
If the score is uninformative or misleading ($\mu_G \le 0$), acceptance stays at least $\exp(-\lvert\mu_G\rvert/T - \log Z_T)$ times that of uniform proposals, a factor that tends to 1 as $T$ grows.
A tempered proposal therefore sits between \srs ($T \to \infty$) and \spgd's fragile top-1 choice ($T \to 0$, which never hits when the top move is not improving), and it pays a bounded price when the gradient is wrong.

How closely may the add ranking be followed?
On ImageNet its top is unreliable.
On Salman RN-18 and Wong RN-50, the top 0.1\% of outside pixels by $\Phi$ has a mean predicted margin change of $-0.07$ to $-0.08$ but a mean true change of $+0.02$ to $+0.05$ (\cref{tab:phibins}): the pixels with the largest gradients are also the most non-linear.
The standardized scores are heavy-tailed, with a kurtosis of 250--410 on the robust ImageNet CNNs against 11--31 on CIFAR-10 (\cref{tab:conc}).
As a result, at $T = 4$ already 18--24\% of the proposal mass on those models falls on the top 1\% of pixels (3--5\% on CIFAR-10), and at $T \le 2$ the proposal effectively chooses among one to three pixels out of 50\,126.
Per step, a colder proposal is still predicted to be accepted more often on most models (\cref{tab:conc}).
Over a full run, however, the coldest add is the worst setting we tried on ImageNet: with Rules 3 and 5 in place, an add at $T = 1$ leaves 13.0\% and 11.0\% of the selection images unbroken, against 7.4\% and 7.8\% at $T = 4$ (\cref{tab:selection}).
A cold add keeps proposing the few top-ranked pixels, so per-step acceptance is not the right criterion, and the add side stays warm: $\Tadd = 4$.
Even at $T = 4$, a proposal that follows the add ranking on both sides is accepted less often than \srs's on all three robust ImageNet CNNs before step 1000 (\cref{fig:diag}, left), which is why the drop side needs a different rule.

\paragraph{Rule 3: choose pixels to drop from a cold ranking.}
The drop score behaves differently (\cref{fig:diag}, right).
Before step 1000 its Spearman correlation with the true loss from removing a pixel is 0.63--0.65 on the four robust ImageNet models and 0.43 on the standard one, against only 0.06--0.12 for the add score at the same states: the ImageNet markers sit high on the vertical axis and close to zero on the horizontal one.
On CIFAR-10 it separates the models by colour: it is at least 0.36 on the five models where \spgd beats \srs and at most 0.25 on the seven where \srs wins, with a rank correlation of 0.94 with the \srs--\spgd gap ($n = 12$, $p = 7 \times 10^{-6}$).
A reliable score over only $k$ pixels is wasted at $T = 4$.
On the four robust ImageNet models the pixel dropped at $T = 1$ costs 35--37\% less margin than a uniformly chosen one, while at $T = 4$ the saving is only 13--14\%, and with a uniform add a cold drop raises the predicted single-swap acceptance by 27--52\% (\cref{tab:conc}).
Each side's temperature should match the reliability of its score: $\Tdrop = 1$.

\paragraph{Rule 4: set values by the sign at add time, and spend no steps on values.}
\begin{proposition}[best-response fill]
\label{prop:fill}
Let $h(v) = \Delta m$ for adding pixel $i$ at $v \in B_i$ have a $\lambda_i$-Lipschitz gradient, let $v^\ast$ be the best-response corner and $u$ a uniformly random corner. Then
$\E_u[h(u)] - h(v^\ast) \ge \norm{g_i}_1 / 2 - \lambda_i C$.
\end{proposition}
Choosing a value needs only the $C$ signs of the gradient at one pixel, whereas choosing a location needs $\Phi$ to order $N$ pixels, each with an error of the size in \cref{lem:reliability}.
The value is the more reliable use of the gradient, and the data agree (\cref{tab:diag}).
Before step 1000 the best-response corner is the true best of the $2^C$ corners in 26--36\% of cases on the RobustBench CIFAR-10 models and 15--23\% on ImageNet (chance is 12.5\%).
With uniform locations, adds at that corner are accepted 1.3--3.0 times as often as adds at a random corner on the RobustBench CIFAR-10 models and 1.06--1.22 times on ImageNet, but not on sAT and sTRADES (0.84--0.90).
Once a pixel's value is set when it is added, further sign steps on all values only compete with swaps for the budget: a variant that spends a quarter of its steps on them is no stronger (\cref{tab:selection}), so every step goes to a swap.

\paragraph{Rule 5: do not propose the same pixel twice while the gradient is unchanged.}
The gradient, and with it both rankings, stays fixed until a move is accepted.
With heavy-tailed scores a tempered proposal keeps returning to the same few pixels between acceptances, and a cold drop over $k$ pixels does the same.
A tabu list~\cite{glover1989tabu} prevents this: every pixel proposed as an add or a drop is excluded until the next acceptance, and the whole list is cleared when either side has fewer than $q$ free pixels.
This restores coverage without warming the proposal.
With the tabu list the tempered add is 1.4 and 0.2 points stronger than a uniform add on the two selection models; without it, it is 1.2 and 1.0 points weaker (\cref{tab:selection}).
On the evaluation points, removing the tabu list costs 1.1--2.0 points on the robust ImageNet CNNs and 4.0 on sTRADES (\cref{sec:ablation}).

\section{The attack: \grs}
\label{sec:method}

\begin{figure}[t]
\centering
\input{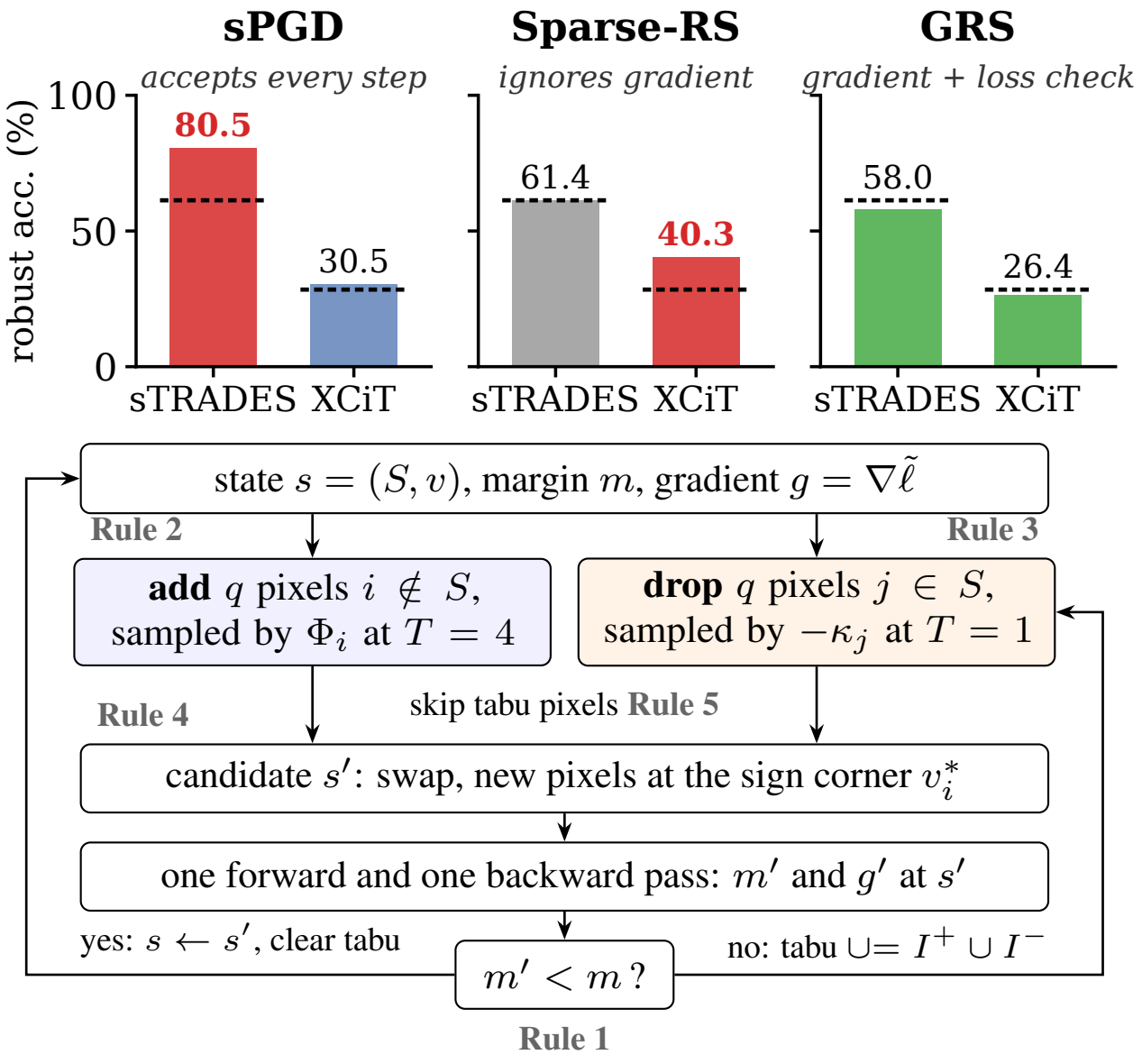}
\caption{Top: robust accuracy (\%, lower is stronger) where the gradient misleads (sTRADES, $k = 20$) and where it helps (ImageNet XCiT-S12, $k = 50$); dashes: \saa; red: over 5 points above it; \spgd: better of its two runs. Bottom: one step of \grs; pixels are drawn $\propto \exp(z/T)$, $z$ the standardized score.}
\label{fig:pipeline}
\end{figure}

\Cref{alg:grs,fig:pipeline} put the five rules together.
The state is a support of $k$ pixels and a value for each of them.
One step costs one forward and one backward pass through the model, the same as \spgd, and the gradient at the accepted state drives the next proposal.
The gradient is that of \spgd's adaptive loss~\cite{zhong2024sparse,croce2020reliable}, written as an objective to decrease: $\tilde\ell = m$ when $m < 0.5$ and $\tilde\ell = -\mathrm{CE}$ (cross-entropy) otherwise, so $g = \nabla m$ near the decision boundary and the scores of \cref{sec:search} apply with $g = \nabla \tilde\ell$.
Scores are standardized over the allowed pixels, and $q$ distinct pixels are drawn in proportion to $\exp(z/T)$ with the Gumbel-top-$q$ trick~\cite{kool2019stochastic}.
The number $q$ follows \srs's schedule: a fraction $p_t$ of $k$ that starts at $p_0 = 0.8$ and halves at fixed iteration marks, rescaled to the iteration budget.
A point stops as soon as it is misclassified, and every success is re-checked with a fresh forward pass.

\begin{algorithm}[t]
\caption{\grs: Gradient-guided Random Search}
\label{alg:grs}
\small
\textbf{Input:} image $x$, label $y$, budget $k$, iterations $n$\\
\textbf{Settings:} $\Tadd = 4$, $\Tdrop = 1$
\begin{enumerate}[leftmargin=1.6em,itemsep=0pt,topsep=2pt,label=\arabic*:]
\item $S \gets k$ random pixels, $v$ random in the box; tabu $\gets \emptyset$
\item $m \gets$ margin and $g \gets \nabla \tilde\ell$ at $x + v_S$
\item \kw{for} $t = 1, \dots, n$ \kw{while} $m \ge 0$:
\item \quad $q \gets \max(1, \mathrm{round}(p_t k))$
\item \quad $\Phi_i \gets \sum_c \max(-g_{ic}(1 - x_{ic}),\, g_{ic} x_{ic})$ for $i \notin S$
\item \quad $\kappa_j \gets -\ip{g_j}{v_j}$ for $j \in S$
\item \quad \kw{if} either side has $< q$ pixels outside tabu: tabu $\gets \emptyset$
\item \quad $I^+ \gets$ $q$ pixels $\notin S \cup$ tabu, $\propto \exp(z(\Phi)/\Tadd)$ \cmt{Rule 2}
\item \quad $I^- \gets$ $q$ pixels $\in S \setminus$ tabu, $\propto \exp(z(-\kappa)/\Tdrop)$ \cmt{Rule 3}
\item \quad tabu $\gets$ tabu $\cup\, I^+ \cup I^-$ \cmt{Rule 5}
\item \quad $S' \gets (S \setminus I^-) \cup I^+$; $v'_i \gets v_i^\ast$ for $i \in I^+$ \cmt{Rule 4}
\item \quad $m' \gets$ margin and $g' \gets \nabla \tilde\ell$ at $x + v'_{S'}$
\item \quad \kw{if} $m' < m$: $S, v, m, g \gets S', v', m', g'$; tabu $\gets \emptyset$ \cmt{Rule 1}
\end{enumerate}
\end{algorithm}

\paragraph{Cost.}
Following~\cite{zhong2024sparse}, we count cost in forward-pass units and count a backward pass as two.
\spgd and \grs therefore cost 3 units per iteration and \srs 1 unit per query, so at 10k iterations \saa costs up to 7 units of 10k forward passes per point and \grs 3.

\paragraph{How the settings were chosen.}
All settings were chosen on points disjoint from the evaluation points (\cref{sec:selection}): $\Tdrop$, the tabu list and swap-only steps on ImageNet validation images 0--499, the rest on CIFAR-10 test points 0--999.
In the ablation, the version with $\Tdrop = 4$, no tabu list and a quarter of its steps spent on sign steps on the values is called \grsz.

\section{Experiments}
\label{sec:experiments}

\paragraph{Models.}
On CIFAR-10~\cite{krizhevsky2009learning} we use a standard PreAct ResNet-18 (PRN-18)~\cite{he2016identity}, the sAT and sTRADES PRN-18 of~\cite{zhong2024sparse} (trained at $k = 120$), and nine RobustBench~\cite{croce2021robustbench} models.
Five of these are $\ell_\infty$-robust: Gowal \etal~\cite{gowal2021improving} PRN-18, Sehwag \etal~\cite{sehwag2022robust} ResNet-18 (RN-18), Cui \etal~\cite{cui2024decoupled} and Wang \etal~\cite{wang2023better} WideResNet-28-10~\cite{zagoruyko2016wide} (WRN-28), and Debenedetti \etal~\cite{debenedetti2023light} XCiT-S12~\cite{elnouby2021xcit}.
Four are $\ell_2$-robust: Rade and Moosavi-Dezfooli~\cite{rade2022reducing} PRN-18, Sehwag \etal RN-18, and Wang \etal and Rebuffi \etal~\cite{rebuffi2021fixing} WRN-28.
On ImageNet~\cite{deng2009imagenet} ($224 \times 224$) we use the standard ResNet-50 (RN-50) and four $\ell_\infty$-robust RobustBench models: Salman \etal~\cite{salman2020adversarially} RN-18 and RN-50, Wong \etal~\cite{wong2020fast} RN-50, and Debenedetti \etal XCiT-S12.

\paragraph{Evaluation setup.}
The evaluation points are CIFAR-10 test images 2000--3999 (1000 of them on the WRN-28 and XCiT models, 2000 otherwise) and ImageNet validation images 1000--1999.
No setting was tuned on them.
The budgets are $k = 5$ and 10 on the standard CIFAR-10 model, $k = 20$ on the robust CIFAR-10 models, and $k = 20$, 50 and 100 on ImageNet, always with 10k iterations or queries per attack.
\spgd (projected and unprojected), \srs, $\sigma$-zero and PGD$_0$ run from their official code with the settings of their own evaluation scripts, except for PGD$_0$'s number of restarts (\cref{sec:settings}).
\saa is the union of the two \spgd runs and \srs, which on every point gives the same result as the official cascade.
All tables report \emph{robust accuracy}, the percentage of all evaluation points that are correctly classified and not broken, so lower means a stronger attack.
Comparisons are paired, with the exact two-sided McNemar test~\cite{mcnemar1947note} on the points that only one of two attacks breaks, and two attacks are \emph{level} when $p \ge 0.05$.
\grs uses seed 0 in the tables and three seeds where stated.

\begin{table*}[t]
\caption{CIFAR-10 at 10k iterations, robust accuracy (\%) on 2000 evaluation points (1000 for WRN-28 and XCiT). Lower is a stronger attack; bold: strongest single attack in the row (unions such as \saa excluded); $b/c$: points broken only by \grs / only by \saa; $^\ast$/$^\dagger$: \grs significantly stronger/weaker than \saa (exact McNemar, $p < 0.05$). \saa is the union of its three members.}
\label{tab:main-cifar}
\centering\footnotesize
\setlength{\tabcolsep}{3pt}
\begin{tabular}{@{}lrrrrrrrrrrr@{}}
\toprule
& & \multicolumn{3}{c}{members of \saa} & & & & & & & \\
\cmidrule(lr){3-5}
model & $k$ & proj & unproj & \srs & \saa & \grs & $b/c$ & \saa$\cup$\grs & $\sigma$-zero & PGD$_0$ & GreedyPixel \\
\midrule
Std.\ PRN-18 & 5 & 8.7 & 18.6 & 6.8 & 5.7 & \textbf{6.0} & 14/20 & 5.0 & 26.7 & 23.5 & 61.1 \\
Std.\ PRN-18 & 10 & 0.2 & 4.0 & \textbf{0.1} & 0.1 & \textbf{0.1} & 1/1 & 0.1 & 3.2 & 6.2 & 36.8 \\
$\ell_\infty$ Gowal PRN-18 & 20 & 9.4 & 15.7 & 7.2 & 6.3 & \textbf{4.6}$^{\ast}$ & 49/15 & 3.8 & 23.3 & 23.7 & 49.3 \\
$\ell_\infty$ Sehwag RN-18 & 20 & 10.0 & 14.6 & 9.2 & 7.6 & \textbf{5.0}$^{\ast}$ & 67/16 & 4.2 & 23.4 & 21.4 & 50.5 \\
$\ell_\infty$ Cui WRN-28 & 20 & 9.1 & 15.2 & 6.5 & 5.6 & \textbf{3.4}$^{\ast}$ & 27/5 & 2.9 & 20.1 & 27.6 & 48.6 \\
$\ell_\infty$ Wang WRN-28 & 20 & 8.7 & 14.2 & 5.8 & 5.2 & \textbf{3.1}$^{\ast}$ & 26/5 & 2.6 & 18.8 & 25.7 & 48.2 \\
$\ell_\infty$ XCiT-S12 & 20 & 9.7 & 14.2 & 16.3 & 8.9 & \textbf{8.5} & 15/11 & 7.4 & 28.3 & 23.4 & 60.9 \\
$\ell_2$ Rade PRN-18 & 20 & 9.3 & 8.5 & 13.3 & 7.5 & \textbf{7.9} & 21/29 & 6.4 & 31.3 & 20.6 & 58.0 \\
$\ell_2$ Sehwag RN-18 & 20 & 6.3 & \textbf{5.8} & 10.9 & 4.7 & 6.4$^{\dagger}$ & 15/50 & 3.9 & 26.9 & 15.1 & 58.4 \\
$\ell_2$ Wang WRN-28 & 20 & 10.7 & 12.4 & 13.5 & 9.0 & \textbf{7.5}$^{\ast}$ & 28/13 & 6.2 & 37.1 & 29.5 & 69.4 \\
$\ell_2$ Rebuffi WRN-28 & 20 & 15.4 & 19.5 & 18.2 & 13.7 & \textbf{11.8}$^{\ast}$ & 26/7 & 11.1 & 35.4 & 24.4 & 63.1 \\
sAT PRN-18 & 20 & 75.4 & 75.0 & 35.2 & 35.2 & \textbf{34.8} & 66/58 & 31.9 & 78.4 & 75.7 & 73.5 \\
sTRADES PRN-18 & 20 & 84.0 & 80.5 & 61.4 & 61.3 & \textbf{58.0}$^{\ast}$ & 100/33 & 56.3 & 85.8 & 85.1 & 83.3 \\
\bottomrule

\end{tabular}
\end{table*}

\begin{table*}[t]
\caption{ImageNet validation images 1000--1999 at 10k iterations, robust accuracy (\%). Lower is a stronger attack; bold, $b/c$ and $^\ast$/$^\dagger$ as in \cref{tab:main-cifar}. 3 seeds: mean $\pm$ sd of \grs. $\sigma$-zero, PGD$_0$ and GreedyPixel at $k = 50$ only.}
\label{tab:main-imagenet}
\centering\footnotesize
\setlength{\tabcolsep}{3pt}
\begin{tabular}{@{}lrrrrrrrrrrrr@{}}
\toprule
& & \multicolumn{3}{c}{members of \saa} & & & & & & & & \\
\cmidrule(lr){3-5}
model & $k$ & proj & unproj & \srs & \saa & \grs & $b/c$ & 3 seeds & \saa$\cup$\grs & $\sigma$-zero & PGD$_0$ & GreedyPixel \\
\midrule
Std.\ RN-50 & 20 & 23.9 & 42.7 & 20.0 & 17.6 & \textbf{17.2} & 26/22 & -- & 15.0 & -- & -- & -- \\
Std.\ RN-50 & 50 & 3.7 & 21.9 & 5.3 & 2.6 & \textbf{3.3} & 6/13 & 3.6$\pm$0.5 & 2.0 & 5.7 & 48.2 & 42.2 \\
Std.\ RN-50 & 100 & \textbf{0.3} & 6.6 & 1.7 & 0.3 & 0.9$^{\dagger}$ & 0/6 & -- & 0.3 & -- & -- & -- \\
Salman RN-18 & 20 & 14.4 & 19.7 & 12.7 & 11.4 & \textbf{10.5} & 19/10 & -- & 9.5 & -- & -- & -- \\
Salman RN-18 & 50 & 5.3 & 7.6 & 6.5 & 4.2 & \textbf{4.0} & 10/8 & 4.2$\pm$0.2 & 3.2 & 5.5 & 26.7 & 19.0 \\
Salman RN-18 & 100 & \textbf{1.3} & 2.2 & 4.1 & 1.3 & \textbf{1.3} & 9/9 & -- & 0.4 & -- & -- & -- \\
Salman RN-50 & 20 & 27.2 & 33.1 & \textbf{19.8} & 19.1 & 20.3 & 22/34 & -- & 16.9 & -- & -- & -- \\
Salman RN-50 & 50 & 10.5 & 16.2 & 9.5 & 7.6 & \textbf{7.7} & 15/16 & 7.9$\pm$0.5 & 6.1 & 8.1 & 36.6 & 34.1 \\
Salman RN-50 & 100 & 4.5 & 6.3 & 5.7 & 3.7 & \textbf{3.0} & 14/7 & -- & 2.3 & -- & -- & -- \\
Wong RN-50 & 20 & 21.4 & 25.7 & 17.9 & 16.9 & \textbf{16.9} & 21/21 & -- & 14.8 & -- & -- & -- \\
Wong RN-50 & 50 & 10.6 & 12.7 & 9.3 & 7.6 & \textbf{6.7} & 15/6 & 7.0$\pm$0.3 & 6.1 & 6.9 & 29.7 & 31.5 \\
Wong RN-50 & 100 & 4.8 & 6.1 & 6.8 & 4.6 & \textbf{3.4}$^{\ast}$ & 17/5 & -- & 2.9 & -- & -- & -- \\
XCiT-S12 & 20 & 48.0 & 50.5 & 51.2 & 46.1 & \textbf{44.6}$^{\ast}$ & 32/17 & -- & 42.9 & -- & -- & -- \\
XCiT-S12 & 50 & 30.5 & 33.3 & 40.3 & 28.4 & \textbf{26.4}$^{\ast}$ & 50/30 & 26.5$\pm$0.2 & 23.4 & 37.2 & 55.1 & 58.6 \\
XCiT-S12 & 100 & 15.9 & 15.8 & 34.0 & 14.7 & \textbf{14.5} & 30/28 & -- & 11.7 & -- & -- & -- \\
\bottomrule

\end{tabular}
\end{table*}

\paragraph{Main result.}
We call the 13 CIFAR-10 configurations and the five ImageNet models at $k = 50$ the 18 \emph{main settings}.
On 17 of them \grs is at least as strong as the best single member of \saa and within one point of \saa itself or stronger (\cref{tab:main-cifar,tab:main-imagenet}).
It is significantly stronger than \saa on 8 settings, seven CIFAR-10 configurations and the ImageNet XCiT, and level with it on 9.
Pooled over CIFAR-10, \grs breaks 455 points that \saa misses and \saa breaks 263 that \grs misses ($p = 7 \times 10^{-13}$); pooled over the four robust ImageNet models the counts are 90 and 60 ($p = 0.018$).
\grs is never weaker than \srs on a main setting, and on ImageNet it is 1.8--13.9 points stronger.
\Cref{fig:examples} (appendix) shows ImageNet images that \grs breaks and neither \spgd proj nor \srs does.

\paragraph{Sparsity.}
At $k = 20$ and 100 (\cref{tab:main-imagenet}) \grs stays within 1.2 points of \saa or stronger, and it is significantly stronger on XCiT at $k = 20$ and on Wong RN-50 at $k = 100$.
A CIFAR-10 sweep over $k = 3$ to 30 with \grsz (\cref{tab:ksweep}) has no significant loss to \saa and three significant wins.

\paragraph{Other $\ell_0$ attacks.}
At the same budget the minimum-norm attack $\sigma$-zero leaves 19--37\% of the points unbroken on the RobustBench CIFAR-10 models, PGD$_0$ 15--30\% and GreedyPixel 48--69\%.
On ImageNet at $k = 50$ PGD$_0$ (27--55\%) and GreedyPixel (19--59\%) stay far behind; \grs is level with $\sigma$-zero on Salman and Wong RN-50 and 1.5--10.8 points stronger on the other three ($p < 0.005$).

\begin{table*}[t]
\caption{CIFAR-10 ablation at 10k steps, robust accuracy (\%). Lower is a stronger attack; bold: best variant in the row; $^\ast$/$^\dagger$: significantly stronger/weaker than the full method of its group (exact McNemar, $p < 0.05$). The \grs variants drop the tabu list or accept every candidate. Each \grsz variant changes one choice of \grsz ($\Tdrop = 4$, no tabu list, value steps on a quarter of its steps); ``no gradient use'' is \srs's proposal inside our code.}
\label{tab:abl-cifar}
\centering\footnotesize
\setlength{\tabcolsep}{2.5pt}
\begin{tabular}{@{}lrrrrrrrrrrr@{}}
\toprule
& & \multicolumn{3}{c}{\grs} & \multicolumn{6}{c}{\grsz} & \\
\cmidrule(lr){3-5}\cmidrule(lr){6-11}
model & $k$ & full & no tabu & accept all & full & uniform prop. & random corner & no value steps & cross-entropy & no gradient use & official \srs \\
\midrule
Std.\ PRN-18 & 5 & 6.0 & 6.1 & 17.5$^{\dagger}$ & \textbf{5.8} & 6.4 & 6.7$^{\dagger}$ & 6.2 & 6.0 & 7.1$^{\dagger}$ & 6.8$^{\dagger}$ \\
$\ell_\infty$ Gowal PRN-18 & 20 & 4.6 & 4.5 & 26.1$^{\dagger}$ & \textbf{4.3} & 4.8 & 5.4$^{\dagger}$ & \textbf{4.3} & 4.4 & 7.5$^{\dagger}$ & 7.2$^{\dagger}$ \\
$\ell_\infty$ Sehwag RN-18 & 20 & \textbf{5.0} & 5.4 & 25.6$^{\dagger}$ & 5.3 & 6.1$^{\dagger}$ & 6.5$^{\dagger}$ & 5.2 & 5.5 & 10.1$^{\dagger}$ & 9.2$^{\dagger}$ \\
$\ell_2$ Sehwag RN-18 & 20 & 6.4 & \textbf{6.1} & 24.4$^{\dagger}$ & 6.3 & 7.6$^{\dagger}$ & 8.8$^{\dagger}$ & 6.3 & 6.7 & 12.2$^{\dagger}$ & 10.9$^{\dagger}$ \\
sTRADES PRN-18 & 20 & \textbf{58.0} & 62.0$^{\dagger}$ & 87.2$^{\dagger}$ & \textbf{58.0} & 59.1$^{\dagger}$ & 60.3$^{\dagger}$ & 58.4 & 58.2 & 61.8$^{\dagger}$ & 61.4$^{\dagger}$ \\
\bottomrule

\end{tabular}
\end{table*}

\paragraph{Equal compute.}
\grs's 3 units are a per-step cost, so \cref{tab:equal} (appendix) gives the search \saa's budget on CIFAR-10.
These runs use \grsz, which is level with \grs on CIFAR-10 (\cref{sec:ablation}).
Run for 23k steps (6.9 units), it is significantly stronger than \saa on 8 of 13 configurations and level on 4.
Two seeds at 10k cost 6 units and beat the single 23k run on 9 of 13 configurations.

\paragraph{Seeds.}
On ImageNet at $k = 50$ the standard deviation of \grs over three seeds is 0.2--0.5 points (\cref{tab:main-imagenet,tab:seeds}).
On CIFAR-10 we have three seeds of \grsz on every configuration, with a standard deviation of 0.03--0.55; \grs has one seed there.

\paragraph{Wall-clock.}
\Cref{tab:timing} (appendix) times the official code and ours on CIFAR-10 on one NVIDIA L40 at batch size 500.
Per 1000 points \grs takes 138--166\,s, against 527--699\,s for the three members of \saa, each run on all points, a factor of 3.7--5.0.
This exceeds the 7/3 unit ratio because the official \srs alone takes 277--423\,s.

\section{Ablation and diagnostics}
\label{sec:ablation}

\paragraph{Components.}
\Cref{tab:abl-cifar} removes one design choice at a time on five CIFAR-10 configurations; \cref{tab:abl-rules-imagenet,tab:abl-imagenet} (appendix) do the same on ImageNet.
Greedy acceptance (Rule 1) has by far the largest effect: accepting every candidate, as \spgd does, leaves 11.5--29.2 points more unbroken on CIFAR-10 and 9.1--24.0 on ImageNet at $k = 50$ ($p < 10^{-27}$), consistent with \cref{prop:accept}.
Removing the tabu list (Rule 5) costs 1.1--2.0 points on the three robust ImageNet CNNs and 4.0 on sTRADES ($p \le 0.03$), and at most 0.4 points elsewhere.
The other CIFAR-10 variants start from \grsz, which is level with \grs on CIFAR-10: pooled over all 13 configurations each breaks exactly 256 points that the other misses.
Replacing the best-response fill with \srs's random corner costs 0.9--2.5 points and is significant on all five configurations; this is the largest effect of a single change to \grsz, consistent with \cref{prop:fill}.
Replacing the tempered add and drop with uniform proposals costs 0.5--1.3 points and is significant on three.
Removing the value steps, or taking the gradient of the cross-entropy instead of the adaptive loss, changes nothing significant.

On ImageNet, removing the cold drop, the tabu list and swap-only steps costs 0.9--2.8 points at $k = 50$ ($p \le 0.002$ on the three robust CNNs and 0.02 on XCiT), 3.5--5.7 points at $k = 20$ on the robust CNNs, and 0.5--1.9 points at $k = 100$ on the robust models.
At $k = 50$ it moves the attack from within 0.7 points of \saa to 1.4--2.4 points above it on the four CNNs, and from below \srs to above it on two of them.
Without the three settings the search leads \srs early but slows down in the late phase, where one or two pixels are swapped per step (\cref{fig:curves}, appendix).

\paragraph{Diagnostics.}
The reliability measurements of \cref{sec:rules} explain most of the differences between models.
The acceptance ratio between the tempered proposal and \srs's (\cref{fig:diag}, left) is 2.9--10.8 before step 100 and 2.1--4.6 before step 1000 on the nine RobustBench CIFAR-10 models, but only 0.6--0.9 on sAT and sTRADES.
These are the two regimes of \cref{prop:temper}.
On the robust ImageNet CNNs the ratio falls to 0.5--0.8 before step 1000.
Splitting the proposal into its parts shows why: there the tempered proposal is accepted 20--55\% \emph{less} often than a uniform proposal with the same best-response fill, while the fill itself still helps (\cref{tab:diag}).
Rules 3 and 5 recover the gradient's value there: the cold drop uses the reliable drop score, and the tabu list keeps the proposal from returning to the non-linear top of the add ranking.

\paragraph{An open point.}
On sAT and sTRADES no single-move statistic favours the gradient.
The add and drop scores are negatively correlated with the truth, the best-response corner is at chance, and the tempered proposal is accepted no more often than \srs's (\cref{tab:diag}).
Yet \grs is level with \srs on sAT and 3.4 points stronger on sTRADES ($b/c = 101/33$, $p = 3 \times 10^{-9}$), and the ablation shows that the tempered proposals, the best-response fill and the tabu list all contribute on sTRADES.

\section{Limitations}
\label{sec:limitations}

The bounds of \cref{sec:search,sec:rules} are per step; we do not estimate $\lambda_i$, and ReLU networks are only piecewise smooth, so the bounds explain the measurements rather than predict them.
\saa breaks significantly more points on the $\ell_2$ Sehwag RN-18 and on the standard RN-50 at $k = 100$, and at $k = 50$ with one of the three seeds (\cref{tab:main-cifar,tab:main-imagenet,tab:seeds}).
Three settings ($\Tdrop$, the tabu list and swap-only steps) were chosen among nine variants on two ImageNet models; their gain on the evaluation images is similar, including on the three other models (\cref{sec:selection}).
Finally, \grs needs gradients, so it cannot replace a black-box attack where they are unavailable.

\section{Conclusion}
A pixel-$\ell_0$ attack is a proposal and an acceptance rule, and how far the gradient can be trusted for each can be measured.
\grs lets the gradient propose and the loss accept, and is within one point of the three-member ensemble or stronger on 17 of 18 settings, and significantly stronger on 8, at 3 instead of up to 7 units.
Code and all per-point results will be released.

\clearpage  

{
    \small
    \bibliographystyle{ieeenat_fullname}
    \bibliography{main}
}

\clearpage
\setcounter{page}{1}
\renewcommand{\thepage}{S\arabic{page}}  
\maketitlesupplementary
\appendix

\section{Hyperparameter selection}
\label{sec:selection}

All settings were chosen on held-out points that are disjoint from the evaluation points of \cref{sec:experiments}.

\paragraph{CIFAR-10.}
Greedy acceptance, the add temperature $\Tadd = 4$, the best-response fill and the schedule were chosen on CIFAR-10 test points 0--999 of three models: the standard model at $k = 5$, and the $\ell_\infty$ Gowal PRN-18 and $\ell_2$ Sehwag RN-18 at $k = 20$.

\paragraph{ImageNet.}
The diagnostics of \cref{sec:rules} on ImageNet validation images 0--499 led us to compare nine settings of the drop temperature, the tabu list, the add temperature and the value steps on the same images, for Salman RN-50 and Wong RN-50 only (\cref{tab:selection}).
We chose the best: cold drop, tabu list and swap-only steps with $\Tadd = 4$, which is \grs.
On the selection images it is 3.2 and 1.8 points stronger than \grsz.
On the evaluation images (\cref{tab:abl-imagenet}) the gains on the same two models are 2.3 and 2.8 points, and on the three models outside the selection comparison they are 1.6 (Salman RN-18), 1.5 (XCiT) and 0.9 points (standard RN-50, $p = 0.15$).
On CIFAR-10 the three settings together have no significant effect on any of the 13 configurations ($p \ge 0.06$, \cref{tab:equal}), and pooled over all 13 each of \grs and \grsz breaks 256 points that the other misses.

\begin{table}[t]
\caption{Selection on ImageNet validation images 0--499 (not used for evaluation), $k = 50$, 10k steps, seed 0, robust accuracy (\%) on Salman RN-50 and Wong RN-50. Lower is a stronger attack; bold: best value in the column. $b/c$: points broken only by the variant / only by the reference (\grsz or the official \srs), pooled over both models; $^\ast$/$^\dagger$: variant significantly stronger/weaker than the reference (exact McNemar, $p < 0.05$). All variants except the first two take no value steps.}
\label{tab:selection}
\centering\scriptsize
\setlength{\tabcolsep}{2.5pt}
\begin{tabular}{@{}lrrll@{}}
\toprule
variant & Salman & Wong & vs \grsz & vs \srs \\
\midrule
\grsz (value steps 25\%) & 10.6 & 9.6 & -- & 32/24 \\
official \srs & 11.0 & 10.8 & 24/32 & -- \\
\srs proposal inside our code & 10.6 & 9.2 & 23/21 & 26/16 \\
no value steps only & 10.6 & 9.4 & 19/18 & 27/18 \\
best-response fill, uniform add/drop & 9.8 & 9.4 & 28/23 & 27/14 \\
cold drop ($\Tdrop = 1$), add at $T = 4$ & 10.0 & 9.2 & 21/16 & 31/18 \\
cold drop, uniform add & 8.8 & 8.2 & 30/14$^\ast$ & 33/9$^\ast$ \\
drop at $T = 0.5$, uniform add & 8.6 & 8.4 & 27/11$^\ast$ & 34/10$^\ast$ \\
tabu, cold drop, uniform add & 8.8 & 8.0 & 27/10$^\ast$ & 30/5$^\ast$ \\
tabu, cold drop, add at $T=4$ (\grs) & \textbf{7.4} & \textbf{7.8} & 28/3$^\ast$ & 38/5$^\ast$ \\
tabu, cold drop, add at $T = 1$ & 13.0 & 11.0 & 20/39$^\dagger$ & 21/32 \\
\bottomrule
\end{tabular}
\end{table}

\section{Settings of the attacks}
\label{sec:settings}

\paragraph{\grs.}
$\Tadd = 4$, $\Tdrop = 1$, and $p_0 = 0.8$ with \srs's halving schedule rescaled to the budget.
New pixels take the best-response corner, proposed pixels are tabu until the next acceptance, acceptance is greedy, and the gradient is that of $\tilde\ell$ (\cref{sec:method}).
The initial support and values are random, and each point stops once it is broken.
Standardized scores use the mean and variance over the allowed pixels of the point.
\grsz differs in three ways: $\Tdrop = 4$, no tabu list, and with probability 0.25 a value step (a sign step of size 0.25 on all values in the support) instead of a swap.

\paragraph{Official attacks.}
\spgd proj, \spgd unproj and \srs run from the \saa repository with the arguments of its evaluation script: $\epsilon = 1$, $\alpha = \beta = 0.25$, patience 3 and the adaptive loss.
\srs uses $p_0 = 0.8$ with cross-entropy on robust models and $p_0 = 0.3$ with the margin loss on standard ones.
$\sigma$-zero runs from its repository for 10k iterations with its defaults (Adam, cosine schedule) and no budget; it minimizes the number of changed features, and a point counts as broken if the returned input differs from the clean one in at most $k$ pixels.
PGD$_0$ runs from the authors' code with $\ell_0$ plus box constraints, 100 steps per restart, step size $120000/255$ and $\lfloor \text{budget} / 100 \rfloor$ restarts.
GreedyPixel runs in its white-box mode (the surrogate is the target model), with its pixel order cut to the top $k$ pixels so that at most $k$ pixels change.
Every success of every attack is re-checked on a fresh forward pass: the point must be misclassified, at most $k$ pixels may differ, and all values must lie in $[0, 1]$.

\section{Additional tables and figures}

\begin{figure*}[t]
\centering
\includegraphics[width=0.92\textwidth]{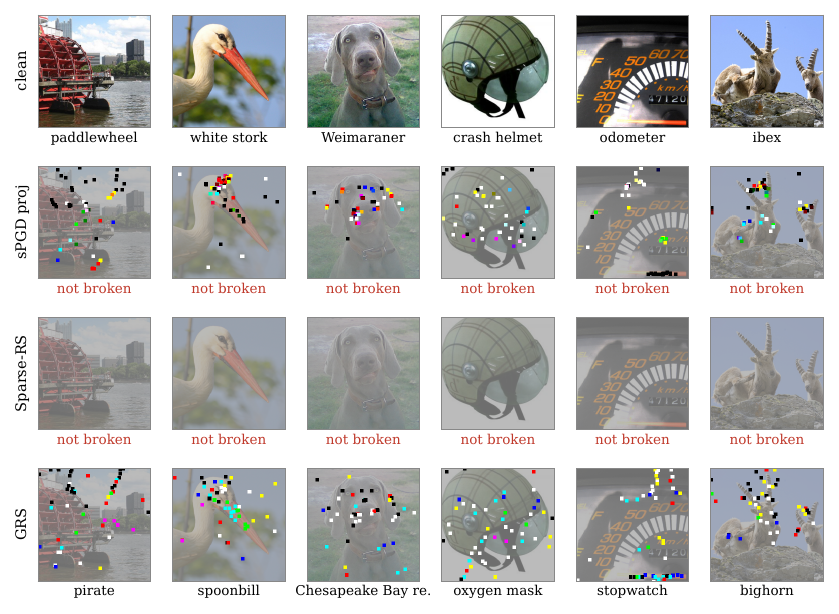}
\caption{Adversarial examples of \spgd proj, \srs and \grs on the Wong \etal RN-50 at $k = 50$ (50 of $224 \times 224$ pixels). Top row: clean image and true class. Other rows: the pixels each attack changes, each drawn as a $7 \times 7$ block of its new colour on a faded copy of the clean image, and the predicted class (red: not broken). \grs breaks all six images, while \spgd proj and \srs break none of them (\srs returns the clean image when it fails).}
\label{fig:examples}
\end{figure*}

\begin{figure*}[t]
\centering
\includegraphics[width=0.8\textwidth]{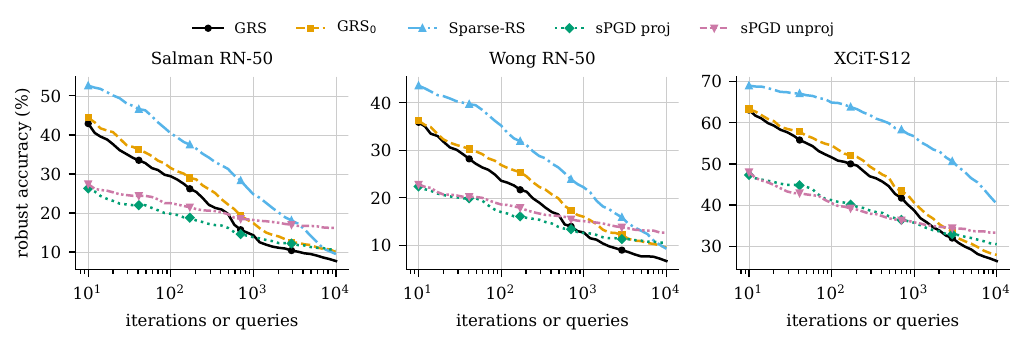}
\caption{Robust accuracy against iterations (\spgd, \grs, \grsz) or queries (\srs) on three ImageNet models at $k = 50$, seed 0. The axis counts steps, not compute: one \grs or \spgd iteration costs 3 units and one \srs query 1. \grsz has no cold drop, no tabu list and spends a quarter of its steps on values.}
\label{fig:curves}
\end{figure*}

\begin{table}[t]
\caption{ImageNet, $k = 50$, robust accuracy (\%) of \grs and \grsz for seeds 0 / 1 / 2, and mean $\pm$ sample sd. Lower is a stronger attack.}
\label{tab:seeds}
\centering\scriptsize
\setlength{\tabcolsep}{2.5pt}
\begin{tabular}{@{}lllll@{}}
\toprule
model & \grs seeds & mean & \grsz seeds & mean \\
\midrule
Std.\ RN-50 & 3.3 / 3.3 / 4.1 & 3.6 $\pm$ 0.5 & 4.2 / 4.0 / 3.4 & 3.9 $\pm$ 0.4 \\
Salman RN-18 & 4.0 / 4.4 / 4.2 & 4.2 $\pm$ 0.2 & 5.6 / 5.3 / 5.3 & 5.4 $\pm$ 0.2 \\
Salman RN-50 & 7.7 / 7.6 / 8.5 & 7.9 $\pm$ 0.5 & 10.0 / 10.5 / 10.7 & 10.4 $\pm$ 0.4 \\
Wong RN-50 & 6.7 / 6.9 / 7.3 & 7.0 $\pm$ 0.3 & 9.5 / 8.2 / 9.0 & 8.9 $\pm$ 0.7 \\
XCiT-S12 & 26.4 / 26.3 / 26.7 & 26.5 $\pm$ 0.2 & 27.9 / 27.5 / 28.6 & 28.0 $\pm$ 0.6 \\
\bottomrule
\end{tabular}
\end{table}

\begin{table}[t]
\caption{ImageNet evaluation images, robust accuracy (\%) of \grs and \grsz, with \srs and \saa for reference. Lower is a stronger attack; bold: strongest of \grs, \grsz and \srs; $b/c$: points broken only by \grs / only by \grsz, with the exact McNemar $p$; $^\ast$/$^\dagger$: \grs significantly stronger/weaker than \grsz ($p < 0.05$).}
\label{tab:abl-imagenet}
\centering\footnotesize
\setlength{\tabcolsep}{2.5pt}
\begin{tabular}{@{}lrrrrrrr@{}}
\toprule
model & $k$ & \grs & \grsz & $b/c$ & $p$ & \srs & \saa \\
\midrule
Std.\ RN-50 & 20 & \textbf{17.2} & \textbf{17.2} & 21/21 & 1.00 & 20.0 & 17.6 \\
Std.\ RN-50 & 50 & \textbf{3.3} & 4.2 & 20/11 & 0.15 & 5.3 & 2.6 \\
Std.\ RN-50 & 100 & \textbf{0.9} & \textbf{0.9} & 5/5 & 1.00 & 1.7 & 0.3 \\
Salman RN-18 & 20 & \textbf{10.5}$^{\ast}$ & 14.0 & 39/4 & $10^{-8}$ & 12.7 & 11.4 \\
Salman RN-18 & 50 & \textbf{4.0}$^{\ast}$ & 5.6 & 18/2 & $10^{-4}$ & 6.5 & 4.2 \\
Salman RN-18 & 100 & \textbf{1.3} & 1.8 & 9/4 & 0.27 & 4.1 & 1.3 \\
Salman RN-50 & 20 & 20.3$^{\ast}$ & 26.0 & 64/7 & $10^{-12}$ & \textbf{19.8} & 19.1 \\
Salman RN-50 & 50 & \textbf{7.7}$^{\ast}$ & 10.0 & 35/12 & 0.0011 & 9.5 & 7.6 \\
Salman RN-50 & 100 & \textbf{3.0}$^{\ast}$ & 4.6 & 22/6 & 0.0037 & 5.7 & 3.7 \\
Wong RN-50 & 20 & \textbf{16.9}$^{\ast}$ & 20.5 & 45/9 & $10^{-7}$ & 17.9 & 16.9 \\
Wong RN-50 & 50 & \textbf{6.7}$^{\ast}$ & 9.5 & 28/0 & $10^{-9}$ & 9.3 & 7.6 \\
Wong RN-50 & 100 & \textbf{3.4}$^{\ast}$ & 5.3 & 21/2 & $10^{-5}$ & 6.8 & 4.6 \\
XCiT-S12 & 20 & \textbf{44.6} & 44.8 & 14/12 & 0.85 & 51.2 & 46.1 \\
XCiT-S12 & 50 & \textbf{26.4}$^{\ast}$ & 27.9 & 26/11 & 0.02 & 40.3 & 28.4 \\
XCiT-S12 & 100 & \textbf{14.5}$^{\ast}$ & 16.0 & 23/8 & 0.011 & 34.0 & 14.7 \\
\bottomrule

\end{tabular}
\end{table}

\begin{table}[t]
\caption{ImageNet evaluation images at $k = 50$, robust accuracy (\%) of \grs, \grs without the tabu list and \grs accepting every candidate, with \srs for reference. Lower is a stronger attack; bold: strongest in the row; $^\ast$/$^\dagger$: variant significantly stronger/weaker than \grs (exact McNemar, $p < 0.05$).}
\label{tab:abl-rules-imagenet}
\centering\footnotesize
\begin{tabular}{@{}lrrrr@{}}
\toprule
model & \grs & no tabu & accept all & \srs \\
\midrule
Std.\ RN-50 & \textbf{3.3} & \textbf{3.3} & 27.3$^{\dagger}$ & 5.3 \\
Salman RN-18 & \textbf{4.0} & 5.1$^{\dagger}$ & 13.1$^{\dagger}$ & 6.5 \\
Salman RN-50 & \textbf{7.7} & 9.1$^{\dagger}$ & 22.9$^{\dagger}$ & 9.5 \\
Wong RN-50 & \textbf{6.7} & 8.7$^{\dagger}$ & 18.9$^{\dagger}$ & 9.3 \\
XCiT-S12 & 26.4 & \textbf{26.3} & 43.7$^{\dagger}$ & 40.3 \\
\bottomrule

\end{tabular}
\end{table}

\begin{table}[t]
\caption{Seconds per 1000 CIFAR-10 points at 10k iterations, one L40, batch 500 (lower is faster). \saa is two \spgd runs plus \srs, with the unprojected run assumed to take as long as the projected one; the ratio is \saa's time over \grs's.}
\label{tab:timing}
\centering\footnotesize
\setlength{\tabcolsep}{3pt}
\begin{tabular}{@{}lrrrrrr@{}}
\toprule
model & $k$ & \spgd proj & \srs & \grs & \saa & ratio \\
\midrule
$\ell_2$ Sehwag RN-18 & 20 & 138 & 423 & 139 & 699 & 5.0$\times$ \\
$\ell_\infty$ Gowal PRN-18 & 20 & 172 & 277 & 166 & 622 & 3.7$\times$ \\
Std.\ PRN-18 & 5 & 117 & 292 & 138 & 527 & 3.8$\times$ \\
\bottomrule

\end{tabular}
\end{table}

\begin{table*}[t]
\caption{Reliability before step 1000 (CIFAR-10 test points 1000--1999, ImageNet validation images 0--499; 64 unbroken points per state): Spearman of the add score with the true add gain (loc), Pearson of the first-order with the true margin change of an add (lin), Spearman of the drop score with the true drop loss (drop), fraction of adds whose best-response corner is the best of the $2^C$ corners (br; chance 0.125), Pearson of the sum of an add and a drop with the swap (add.), and acceptance ratios of three proposals: tempered over \srs before steps 100 and 1000, uniform locations with best-response fill over \srs (fill), and tempered over uniform locations with the same fill (temp.).}
\label{tab:diag}
\centering\footnotesize
\setlength{\tabcolsep}{2.5pt}
\begin{tabular}{@{}lrrrrrrrrr@{}}
\toprule
model & loc & lin & drop & br & add. & ratio$_{100}$ & ratio$_{1000}$ & fill & temp. \\
\midrule
Std.\ PRN-18 & 0.00 & -0.09 & 0.14 & 0.23 & 0.95 & 1.8 & 1.0 & 1.21 & 0.79 \\
$\ell_\infty$ Gowal PRN-18 & 0.28 & 0.19 & 0.17 & 0.31 & 0.99 & 4.5 & 2.2 & 1.48 & 1.50 \\
$\ell_\infty$ Sehwag RN-18 & 0.26 & 0.19 & 0.25 & 0.33 & 0.99 & 4.7 & 2.3 & 2.14 & 1.09 \\
$\ell_\infty$ Cui WRN-28 & 0.22 & 0.17 & 0.07 & 0.29 & 0.99 & 2.9 & 2.5 & 1.82 & 1.36 \\
$\ell_\infty$ Wang WRN-28 & 0.24 & 0.19 & 0.07 & 0.29 & 0.99 & 4.0 & 2.4 & 1.60 & 1.53 \\
$\ell_\infty$ XCiT-S12 & 0.24 & 0.18 & 0.37 & 0.26 & 0.98 & 3.8 & 2.1 & 1.29 & 1.66 \\
$\ell_2$ Rade PRN-18 & 0.27 & 0.28 & 0.44 & 0.35 & 0.99 & 10.2 & 4.6 & 2.97 & 1.54 \\
$\ell_2$ Sehwag RN-18 & 0.21 & 0.14 & 0.40 & 0.36 & 0.99 & 10.8 & 4.1 & 2.61 & 1.57 \\
$\ell_2$ Wang WRN-28 & 0.22 & 0.19 & 0.39 & 0.31 & 0.99 & 8.4 & 3.9 & 2.50 & 1.55 \\
$\ell_2$ Rebuffi WRN-28 & 0.11 & 0.10 & 0.36 & 0.30 & 0.99 & 5.4 & 2.1 & 1.31 & 1.58 \\
sAT PRN-18 & -0.24 & -0.29 & -0.15 & 0.13 & 1.00 & 0.6 & 0.9 & 0.90 & 1.04 \\
sTRADES PRN-18 & -0.17 & -0.13 & -0.14 & 0.11 & 1.00 & 0.6 & 0.8 & 0.84 & 1.00 \\
Std.\ RN-50 & -0.17 & -0.21 & 0.43 & 0.21 & 0.99 & 1.6 & 1.3 & 1.20 & 1.10 \\
Salman RN-18 & 0.07 & -0.05 & 0.64 & 0.18 & 1.00 & 1.8 & 0.5 & 1.10 & 0.45 \\
Salman RN-50 & 0.06 & 0.01 & 0.65 & 0.15 & 1.00 & 2.2 & 0.8 & 1.06 & 0.80 \\
Wong RN-50 & 0.12 & -0.01 & 0.64 & 0.19 & 1.00 & 1.2 & 0.6 & 1.22 & 0.48 \\
XCiT-S12 & 0.12 & 0.04 & 0.63 & 0.23 & 1.00 & 4.3 & 1.6 & 1.16 & 1.37 \\
\bottomrule

\end{tabular}
\end{table*}

\begin{table*}[t]
\caption{Proposal concentration and drop quality before step 1000 (same states as \cref{tab:diag}). Kurtosis of the standardized add score; percentage of the $T = 4$ add proposal's mass on the top 1\% of outside pixels; effective number of outside pixels (perplexity) of the $T = 2$ add proposal; single-swap acceptance predicted from additivity with both sides at $T = 1$ / $T = 4$ / uniform; mean margin increase caused by the dropped pixel at $T = 1$ / $T = 4$ / uniform (lower is better); predicted single-swap acceptance with a uniform add and a uniform / $T = 1$ drop. --: not computed.}
\label{tab:conc}
\centering\scriptsize
\setlength{\tabcolsep}{3pt}
\begin{tabular}{@{}lrrrccc@{}}
\toprule
model & kurtosis & top 1\% (\%) & pixels at $T=2$ & acceptance $T{=}1$ / 4 / unif. & drop cost $T{=}1$ / 4 / unif. & uniform add: unif. / cold drop \\
\midrule
Std.\ PRN-18 & 20 & 4.1 & 625 & 0.004 / 0.002 / 0.002 & 2.307 / 2.356 / 2.380 & -- \\
$\ell_\infty$ Gowal PRN-18 & 28 & 4.6 & 521 & 0.086 / 0.018 / 0.012 & 0.263 / 0.263 / 0.267 & 0.012 / 0.015 \\
$\ell_\infty$ Sehwag RN-18 & 24 & 4.7 & 492 & 0.082 / 0.021 / 0.014 & 0.190 / 0.203 / 0.210 & -- \\
$\ell_\infty$ Cui WRN-28 & 31 & 4.8 & 503 & 0.066 / 0.023 / 0.018 & 0.235 / 0.231 / 0.232 & -- \\
$\ell_\infty$ Wang WRN-28 & 31 & 4.5 & 533 & 0.089 / 0.020 / 0.015 & 0.237 / 0.240 / 0.243 & -- \\
$\ell_\infty$ XCiT-S12 & 19 & 3.8 & 671 & 0.077 / 0.018 / 0.012 & 0.141 / 0.154 / 0.161 & -- \\
$\ell_2$ Rade PRN-18 & 11 & 3.3 & 737 & 0.125 / 0.019 / 0.011 & 0.238 / 0.253 / 0.263 & -- \\
$\ell_2$ Sehwag RN-18 & 13 & 3.4 & 718 & 0.100 / 0.015 / 0.009 & 0.323 / 0.349 / 0.362 & -- \\
$\ell_2$ Wang WRN-28 & 11 & 3.2 & 739 & 0.098 / 0.022 / 0.014 & 0.231 / 0.249 / 0.258 & -- \\
$\ell_2$ Rebuffi WRN-28 & 14 & 3.4 & 725 & 0.101 / 0.024 / 0.017 & 0.195 / 0.204 / 0.212 & -- \\
sAT PRN-18 & 22 & 4.2 & 608 & 0.011 / 0.027 / 0.028 & 0.438 / 0.337 / 0.322 & -- \\
sTRADES PRN-18 & 14 & 3.5 & 715 & 0.029 / 0.027 / 0.026 & 0.336 / 0.237 / 0.221 & 0.026 / 0.013 \\
Std.\ RN-50 & 55 & 5.5 & 1018 & 0.031 / 0.051 / 0.052 & 0.111 / 0.124 / 0.133 & 0.052 / 0.071 \\
Salman RN-18 & 410 & 24.0 & 2 & 0.235 / 0.147 / 0.156 & 0.032 / 0.044 / 0.051 & 0.156 / 0.198 \\
Salman RN-50 & 303 & 18.5 & 2 & -- / 0.150 / 0.126 & 0.037 / 0.050 / 0.058 & 0.126 / 0.185 \\
Wong RN-50 & 250 & 21.5 & 2 & 0.206 / 0.131 / 0.137 & 0.038 / 0.052 / 0.061 & 0.137 / 0.198 \\
XCiT-S12 & 128 & 8.7 & 20 & 0.259 / 0.112 / 0.091 & 0.021 / 0.028 / 0.033 & 0.091 / 0.138 \\
\bottomrule

\end{tabular}
\end{table*}

\begin{table}[t]
\caption{Outside pixels binned by the rank of their add score $\Phi$, at the states of \cref{tab:diag} before step 1000: mean predicted and true margin change of adding the pixel at its best-response corner (negative is good).}
\label{tab:phibins}
\centering\footnotesize
\setlength{\tabcolsep}{3pt}
\begin{tabular}{@{}llrr@{}}
\toprule
model & $\Phi$ bin & predicted & true \\
\midrule
Salman RN-18 & top 0.1\% & $-0.074$ & $+0.020$ \\
Salman RN-18 & 0.1--1\% & $-0.029$ & $+0.012$ \\
Salman RN-18 & 1--10\% & $-0.007$ & $+0.016$ \\
Salman RN-18 & rest & $-0.000$ & $+0.003$ \\
Wong RN-50 & top 0.1\% & $-0.079$ & $+0.053$ \\
Wong RN-50 & rest & $-0.001$ & $+0.006$ \\
$\ell_\infty$ Gowal PRN-18 & top 0.1\% & $-0.461$ & $-0.072$ \\
$\ell_\infty$ Gowal PRN-18 & rest & $-0.019$ & $+0.005$ \\
$\ell_2$ Sehwag RN-18 & top 0.1\% & $-0.564$ & $-0.110$ \\
$\ell_2$ Sehwag RN-18 & rest & $-0.049$ & $-0.013$ \\
\bottomrule
\end{tabular}
\end{table}

\begin{table*}[t]
\caption{CIFAR-10 at equal compute, robust accuracy (\%), with the cost in units of 10k forward passes per point in brackets. Lower is a stronger attack; no entry is bold because the costs differ. The two-seed and 23k-step runs use \grsz, which is level with \grs on CIFAR-10. $b/c$: points broken only by the 23k run / only by \saa, with the exact McNemar $p$; $^\ast$/$^\dagger$: 23k run significantly stronger/weaker than \saa ($p < 0.05$).}
\label{tab:equal}
\centering\footnotesize
\setlength{\tabcolsep}{2.5pt}
\begin{tabular}{@{}lrrrrrrrrr@{}}
\toprule
model & $k$ & \saa [7] & \grs [3] & \grsz [3] & \grsz 2 seeds [6] & \grsz 23k [6.9] & proj$\cup$unproj [6] & $b/c$ & $p$ \\
\midrule
Std.\ PRN-18 & 5 & 5.7 & 6.0 & 5.8 & 4.6 & 5.2 & 8.6 & 22/12 & 0.12 \\
Std.\ PRN-18 & 10 & 0.1 & 0.1 & 0.1 & 0.1 & 0.1 & 0.2 & 1/0 & 1.00 \\
$\ell_\infty$ Gowal PRN-18 & 20 & 6.3 & 4.6 & 4.3 & 3.3 & 3.6$^{\ast}$ & 9.2 & 61/8 & $10^{-11}$ \\
$\ell_\infty$ Sehwag RN-18 & 20 & 7.6 & 5.0 & 5.3 & 4.1 & 4.4$^{\ast}$ & 9.7 & 75/12 & $10^{-12}$ \\
$\ell_\infty$ Cui WRN-28 & 20 & 5.6 & 3.4 & 3.3 & 2.4 & 2.9$^{\ast}$ & 9.0 & 28/1 & $10^{-7}$ \\
$\ell_\infty$ Wang WRN-28 & 20 & 5.2 & 3.1 & 3.2 & 2.8 & 2.5$^{\ast}$ & 8.7 & 29/2 & $10^{-7}$ \\
$\ell_\infty$ XCiT-S12 & 20 & 8.9 & 8.5 & 8.9 & 7.5 & 7.6$^{\ast}$ & 9.2 & 23/10 & 0.035 \\
$\ell_2$ Rade PRN-18 & 20 & 7.5 & 7.9 & 8.1 & 6.6 & 7.3 & 7.6 & 33/29 & 0.70 \\
$\ell_2$ Sehwag RN-18 & 20 & 4.7 & 6.4 & 6.3 & 5.1 & 5.6$^{\dagger}$ & 4.9 & 17/35 & 0.018 \\
$\ell_2$ Wang WRN-28 & 20 & 9.0 & 7.5 & 8.5 & 6.4 & 7.8 & 9.7 & 28/16 & 0.096 \\
$\ell_2$ Rebuffi WRN-28 & 20 & 13.7 & 11.8 & 11.9 & 10.6 & 10.7$^{\ast}$ & 14.8 & 34/4 & $10^{-7}$ \\
sAT PRN-18 & 20 & 35.2 & 34.8 & 34.2 & 32.4 & 32.1$^{\ast}$ & 73.6 & 91/30 & $10^{-8}$ \\
sTRADES PRN-18 & 20 & 61.3 & 58.0 & 58.0 & 55.8 & 55.3$^{\ast}$ & 80.3 & 138/17 & $10^{-25}$ \\
\bottomrule

\end{tabular}
\end{table*}

\begin{table*}[t]
\caption{CIFAR-10 sparsity sweep with \grsz ($\Tdrop = 4$, no tabu list, value steps; 1000 evaluation points per row), robust accuracy (\%). Lower is a stronger attack; bold: strongest single attack in the row; $b/c$: points broken only by \grsz / only by \saa, with the exact McNemar $p$; $^\ast$/$^\dagger$: \grsz significantly stronger/weaker than \saa ($p < 0.05$).}
\label{tab:ksweep}
\centering\footnotesize
\setlength{\tabcolsep}{2.5pt}
\begin{tabular}{@{}lrrrrrrrr@{}}
\toprule
model & $k$ & proj & unproj & \srs & \saa & \grsz & $b/c$ & $p$ \\
\midrule
Std.\ PRN-18 & 3 & 27.6 & 36.1 & 22.0 & 20.6 & \textbf{21.8} & 15/27 & 0.088 \\
Std.\ PRN-18 & 8 & 1.3 & 7.6 & 1.0 & 0.6 & \textbf{0.7} & 1/2 & 1.00 \\
$\ell_\infty$ Gowal PRN-18 & 5 & 52.8 & 57.5 & 45.5 & 44.3 & \textbf{45.2} & 20/29 & 0.25 \\
$\ell_\infty$ Gowal PRN-18 & 10 & 31.3 & 36.4 & 24.9 & 23.6 & \textbf{22.5} & 28/17 & 0.14 \\
$\ell_\infty$ Gowal PRN-18 & 15 & 17.6 & 24.9 & 11.4 & 10.8 & \textbf{9.5}$^{\ast}$ & 25/12 & 0.047 \\
$\ell_\infty$ Gowal PRN-18 & 30 & 1.3 & 4.3 & 1.0 & 0.7 & \textbf{0.4} & 4/1 & 0.38 \\
$\ell_2$ Sehwag RN-18 & 5 & 62.0 & 61.2 & 63.6 & 59.7 & \textbf{59.6} & 19/18 & 1.00 \\
$\ell_2$ Sehwag RN-18 & 10 & 34.4 & 33.0 & 38.1 & 31.3 & \textbf{32.1} & 22/30 & 0.33 \\
$\ell_2$ Sehwag RN-18 & 15 & 15.7 & 14.8 & 20.7 & 13.2 & \textbf{14.3} & 17/28 & 0.14 \\
$\ell_2$ Sehwag RN-18 & 30 & 0.4 & \textbf{0.2} & 2.2 & 0.0 & 0.5 & 0/5 & 0.062 \\
sAT PRN-18 & 5 & 80.4 & 80.1 & 73.0 & 72.9 & \textbf{71.2}$^{\ast}$ & 24/7 & 0.0033 \\
sAT PRN-18 & 10 & 78.6 & 78.1 & 59.2 & 59.2 & \textbf{57.3}$^{\ast}$ & 28/9 & 0.0026 \\
sAT PRN-18 & 15 & 77.8 & 77.2 & 47.2 & 47.2 & \textbf{45.9} & 36/23 & 0.12 \\
sAT PRN-18 & 30 & 72.0 & 73.9 & \textbf{19.5} & 19.5 & 19.7 & 28/30 & 0.90 \\
\bottomrule

\end{tabular}
\end{table*}

\clearpage
\onecolumn
\section{Proofs}
\label{sec:proofs}

\paragraph{Notation and assumption.}
Fix a state $s = (S, v)$ with adversarial input $x'$, a pixel $i \notin S$ and a value $v \in B_i$.
Write $e_i(v) \in \R^{C \times N}$ for the perturbation that equals $v$ at pixel $i$ and $0$ elsewhere, so that adding pixel $i$ at value $v$ moves the input from $x'$ to $x' + e_i(v)$.
Since $i \notin S$, pixel $i$ of $x'$ equals $x_i$, and the new pixel $x_i + v$ lies in $[0, 1]^C$.
Write $\nabla_i m(y) \in \R^C$ for the part of the gradient that belongs to pixel $i$, so $g_i = \nabla_i m(x')$.
We assume that $m$ is differentiable along every such segment and that $\nabla_i m$ is $\lambda_i$-Lipschitz in pixel $i$'s coordinates:
\begin{equation}
  \norm{\nabla_i m\big(x' + e_i(v)\big) - \nabla_i m\big(x' + e_i(v')\big)}_2 \;\le\; \lambda_i \norm{v - v'}_2
  \qquad \text{for all } v, v' \in B_i .
  \label{eq:lambda}
\end{equation}
ReLU networks satisfy this only piecewise, so $\lambda_i$ is a local smoothness proxy.
For every $v \in B_i$ each coordinate satisfies $v_c \in [-x_{ic}, 1 - x_{ic}] \subseteq [-1, 1]$, hence
\begin{equation}
  \norm{v}_2^2 \;=\; \sum_{c=1}^{C} v_c^2 \;\le\; C .
  \label{eq:vnorm}
\end{equation}

\subsection{The add score and the best-response corner}

\begin{lemma}[closed form of the add score]
\label{lem:phi}
$\Phi_i = \sum_{c} \max\big(-g_{ic}(1 - x_{ic}),\; g_{ic}\, x_{ic}\big)$, and the maximum defining $\Phi_i$ is attained at the best-response corner $v_i^\ast$, whose channel $c$ is $1 - x_{ic}$ if $g_{ic} < 0$ and $-x_{ic}$ otherwise.
\end{lemma}
\begin{proof}
\begin{align}
\Phi_i &= \max_{v \in B_i} \ip{-g_i}{v}
  && \text{(Definition of $\Phi_i$)} \\
&= \max_{v \in B_i} \sum_{c=1}^{C} (-g_{ic})\, v_c
  && \text{(Inner product as a sum over channels)} \\
&= \sum_{c=1}^{C} \; \max_{v_c \in [-x_{ic},\, 1 - x_{ic}]} (-g_{ic})\, v_c
  && \text{($B_i$ is a product of intervals; the objective is separable)} \\
&= \sum_{c=1}^{C} \max\big(-g_{ic}(1 - x_{ic}),\; g_{ic}\, x_{ic}\big)
  && \text{(A linear function on an interval peaks at an endpoint)}
\end{align}
The endpoint that attains the maximum in channel $c$ is $1 - x_{ic}$ when $-g_{ic} > 0$ and $-x_{ic}$ when $-g_{ic} \le 0$, which is $v_i^\ast$.
\end{proof}

\subsection{Proof of \cref{lem:reliability} (first-order reliability)}

Let $\varphi(t) = m\big(x' + t\, e_i(v)\big)$ for $t \in [0, 1]$, so that $\varphi(0) = m(x')$ and $\varphi(1)$ is the margin after the add.
\begin{align}
\Delta m - \ip{g_i}{v} &= \varphi(1) - \varphi(0) - \ip{g_i}{v}
  && \text{(Definition of $\Delta m$)} \\
&= \int_0^1 \varphi'(t)\, dt - \ip{g_i}{v}
  && \text{(Fundamental theorem of calculus)} \\
&= \int_0^1 \ip{\nabla_i m\big(x' + t\, e_i(v)\big)}{v}\, dt - \ip{g_i}{v}
  && \text{(Chain rule; $e_i(v)$ is zero outside pixel $i$)} \\
&= \int_0^1 \ip{\nabla_i m\big(x' + t\, e_i(v)\big) - \nabla_i m(x')}{v}\, dt
  && \text{($g_i = \nabla_i m(x')$ and $\textstyle\int_0^1 dt = 1$)}
  \label{eq:taylor}
\end{align}
Bounding the integrand gives
\begin{align}
\big\lvert \Delta m - \ip{g_i}{v} \big\rvert
&\le \int_0^1 \big\lvert \ip{\nabla_i m\big(x' + t\, e_i(v)\big) - \nabla_i m(x')}{v} \big\rvert\, dt
  && \text{(Triangle inequality for integrals, applied to \cref{eq:taylor})} \\
&\le \int_0^1 \norm{\nabla_i m\big(x' + t\, e_i(v)\big) - \nabla_i m(x')}_2\, \norm{v}_2\, dt
  && \text{(Cauchy--Schwarz)} \\
&\le \int_0^1 \lambda_i\, \norm{t\, v}_2\, \norm{v}_2\, dt
  && \text{(\Cref{eq:lambda} with $v' = 0$; $x' + e_i(0) = x'$)} \\
&= \lambda_i\, \norm{v}_2^2 \int_0^1 t\, dt
  && \text{(Pull out constants)} \\
&= \frac{\lambda_i\, \norm{v}_2^2}{2}
  \;\le\; \frac{\lambda_i\, C}{2}
  && \text{($\textstyle\int_0^1 t\, dt = 1/2$; \cref{eq:vnorm})}
\end{align}
which is the claim. \qed

\paragraph{The two consequences stated after \cref{lem:reliability}.}
At the best-response corner, \cref{lem:phi} gives $\ip{g_i}{v_i^\ast} = -\Phi_i$, so
\begin{align}
\Delta m(v_i^\ast)
&\le \ip{g_i}{v_i^\ast} + \frac{\lambda_i C}{2}
  && \text{(Upper half of \cref{lem:reliability})} \\
&= -\Phi_i + \frac{\lambda_i C}{2}
  && \text{(\cref{lem:phi})} \\
&< 0 \quad \text{whenever } \Phi_i > \frac{\lambda_i C}{2} .
  && \text{(Assumption)}
\end{align}
For two pixels $i, i' \notin S$, each added at its best-response corner,
\begin{align}
\Delta m_i
&\le -\Phi_i + \frac{\lambda_i C}{2}
  && \text{(As above, for pixel $i$)} \\
&< -\Phi_{i'} - \frac{\lambda_{i'} C}{2}
  && \text{(Assume $\Phi_i - \Phi_{i'} > (\lambda_i + \lambda_{i'})\, C / 2$)} \\
&\le \Delta m_{i'}
  && \text{(Lower half of \cref{lem:reliability} for pixel $i'$)}
\end{align}
so the first-order order of the two pixels is then guaranteed to be the true order. \qed

\subsection{Proof of \cref{prop:accept} (accepting every step)}

\paragraph{Part (a).}
\Cref{lem:reliability} certifies that an add at value $v$ decreases $m$ only if its upper bound $\ip{g_i}{v} + \lambda_i C / 2$ is negative.
The smallest such bound over all values of pixel $i$ is
\begin{align}
\min_{v \in B_i} \Big( \ip{g_i}{v} + \frac{\lambda_i C}{2} \Big)
&= -\max_{v \in B_i} \ip{-g_i}{v} + \frac{\lambda_i C}{2}
  && \text{($\lambda_i C / 2$ does not depend on $v$)} \\
&= -\Phi_i + \frac{\lambda_i C}{2}
  && \text{(Definition of $\Phi_i$)} \\
&\ge 0
  && \text{(Assumption $\Phi_i \le \lambda_i C / 2$)}
\end{align}
This holds for every pixel $i$, so no single-pixel add carries a certified decrease of $m$, and an attack that accepts every iterate gets no descent certificate from \cref{lem:reliability}. \qed

\paragraph{Part (b).}
Condition on the support $S_t$ evaluated at iteration $t$; by assumption it is independent of $\mathcal{H}$, and $\mathcal{H}$ is uniform over the $\binom{N}{\gamma}$ subsets of size $\gamma$.
\begin{align}
\Prob\big(\lvert S_t \cap \mathcal{H} \rvert \ge r \,\big|\, S_t\big)
&= \Prob\big(\exists R \subseteq S_t,\ \lvert R \rvert = r,\ R \subseteq \mathcal{H} \,\big|\, S_t\big)
  && \text{(At least $r$ helpful pixels iff some $r$-subset is helpful)} \\
&\le \sum_{R \subseteq S_t,\ \lvert R \rvert = r} \Prob\big(R \subseteq \mathcal{H}\big)
  && \text{(Union bound; $S_t$ independent of $\mathcal{H}$)} \\
&= \binom{k}{r} \binom{N - r}{\gamma - r} \Big/ \binom{N}{\gamma}
  && \text{($\lvert S_t \rvert = k$; count the $\gamma$-subsets containing $R$)} \\
&= \binom{k}{r} \prod_{l=0}^{r-1} \frac{\gamma - l}{N - l}
  && \text{(Expand the binomial coefficients)} \\
&\le \binom{k}{r} \Big(\frac{\gamma}{N}\Big)^{r}
  && \text{($\tfrac{\gamma - l}{N - l} \le \tfrac{\gamma}{N}$ because $\gamma \le N$)}
\end{align}
The bound does not depend on $S_t$, so it also holds unconditionally, and a union bound over $t$ iterations multiplies it by $t$. \qed

\paragraph{The numerical example.}
For $N = 1024$, $k = 20$, $\gamma = 50$ and $r = 10$,
\begin{equation}
\binom{20}{10} \Big(\frac{50}{1024}\Big)^{10} \;=\; 184\,756 \times 7.7 \times 10^{-14} \;\approx\; 1.4 \times 10^{-8} ,
\end{equation}
so a success needs on the order of $1 / (1.4 \times 10^{-8}) \approx 7 \times 10^{7}$ iterations, the number quoted in \cref{sec:search}.

\subsection{Proof of \cref{prop:blind} (greedy acceptance of blind proposals)}

Suppose the current support holds $j < r$ helpful pixels.
A single swap drops a pixel $i^-$ drawn uniformly from $S$ and adds a pixel $i^+$ drawn uniformly from the $N - k$ pixels outside $S$, independently.
Under the additive model the swap changes the margin by
\begin{equation}
\Delta m \;=\; -a\,\big(\mathbb{1}[i^+ \in \mathcal{H}] - \mathbb{1}[i^- \in \mathcal{H}]\big),
\end{equation}
which is negative exactly when $i^+ \in \mathcal{H}$ and $i^- \notin \mathcal{H}$.
Let $p_j$ be the probability that one step is accepted at level $j$.
\begin{align}
p_j &= \Prob\big(i^- \notin \mathcal{H}\big)\, \Prob\big(i^+ \in \mathcal{H}\big)
  && \text{(Greedy: accept iff $\Delta m < 0$; $i^-$ and $i^+$ independent)} \\
&= \frac{k - j}{k} \cdot \frac{\gamma - j}{N - k}
  && \text{($S$ holds $k - j$ unhelpful pixels; $\gamma - j$ helpful pixels lie outside $S$)}
\end{align}
An accepted swap raises the level from $j$ to exactly $j + 1$, and a rejected swap leaves the state unchanged.
Proposals are drawn afresh at every step, so the time $\tau_j$ spent at level $j$ is geometric with success probability $p_j$.
Starting from level $j_0$,
\begin{align}
\E[\tau] &= \sum_{j = j_0}^{r - 1} \E[\tau_j]
  && \text{(The levels $j_0, \dots, r - 1$ are visited in order)} \\
&= \sum_{j = j_0}^{r - 1} \frac{1}{p_j}
  && \text{(Mean of a geometric distribution)} \\
&= \sum_{j = j_0}^{r - 1} \frac{k\,(N - k)}{(k - j)(\gamma - j)}
  && \text{(Substitute $p_j$)} \\
&\le (r - j_0)\, \frac{k\,(N - k)}{(k - r + 1)(\gamma - r + 1)}
  && \text{(Each term is largest at $j = r - 1$)} \\
&\le \frac{r\,k\,(N - k)}{(k - r + 1)(\gamma - r + 1)}
  && \text{($j_0 \ge 0$)}
\end{align}
None of these quantities involves the gradient, because the proposals do not. \qed

\paragraph{The numerical example.}
For $N = 1024$, $k = 20$, $\gamma = 50$, $r = 10$ and $j_0 = 0$ the exact sum is
\begin{equation}
\sum_{j=0}^{9} \frac{20 \times 1004}{(20 - j)(50 - j)} \;=\; 20.1 + 21.6 + 23.2 + 25.1 + 27.3 + 29.7 + 32.6 + 35.9 + 39.8 + 44.5 \;\approx\; 300 .
\end{equation}

\subsection{Proof of \cref{prop:temper} (tempered proposals)}

Let $\mathcal{M}$ be the finite set of moves, $U$ the uniform distribution on it, and $Q(\omega) = U(\omega)\, w(\omega)$ with $w = e^{z/T} / Z_T$ and $Z_T = \E_U[e^{z/T}]$, so that $Q$ sums to one.

\paragraph{The identity.}
\begin{align}
\Prob_Q(\text{accept}) &= Q(G)
  && \text{(Greedy: a proposal is accepted iff it improves)} \\
&= \sum_{\omega \in G} U(\omega)\, w(\omega)
  && \text{(Definition of $Q$)} \\
&= U(G) \sum_{\omega \in G} \frac{U(\omega)}{U(G)}\, w(\omega)
  && \text{(Multiply and divide by $U(G)$)} \\
&= \Prob_U(G)\; \E_U[w \mid G]
  && \text{(Definition of conditional expectation)}
\end{align}

\paragraph{The lower bound.}
\begin{align}
\frac{\Prob_Q(G)}{\Prob_U(G)} &= \E_U\big[\exp\big(z/T - \log Z_T\big) \,\big|\, G\big]
  && \text{(Identity above; definition of $w$)} \\
&\ge \exp\Big(\E_U\big[z/T - \log Z_T \,\big|\, G\big]\Big)
  && \text{(Jensen's inequality; $\exp$ is convex)} \\
&= \exp\Big(\frac{\E_U[z \mid G]}{T} - \log Z_T\Big)
  && \text{(Linearity of expectation)}
\end{align}

\paragraph{The upper bound.}
\begin{align}
\frac{\Prob_Q(G)}{\Prob_U(G)} &= \E_U[w \mid G]
  && \text{(Identity above)} \\
&\le \max_{\omega \in G} w(\omega)
  && \text{(An average is at most its largest term)} \\
&= \exp\Big(\frac{\max_{G} z}{T} - \log Z_T\Big)
  && \text{($w$ is increasing in $z$)}
\end{align}
\qed

\paragraph{The normalizer is at least one.}
\begin{align}
Z_T = \E_U\big[e^{z/T}\big]
&\ge e^{\E_U[z]/T}
  && \text{(Jensen's inequality)} \\
&= e^{0} = 1
  && \text{($z$ is standardized under $U$)}
\end{align}
so $\log Z_T \ge 0$. Equality in Jensen's inequality would require $z$ to be constant under $U$; $z$ has variance 1, so $Z_T > 1$ and $\log Z_T > 0$ for every finite $T$. If $z \sim \mathcal{N}(0, 1)$ under $U$, then $Z_T = \E[e^{z/T}] = e^{1/(2T^2)}$ by the moment generating function of the standard normal, so $\log Z_T = 1/(2T^2)$.

\paragraph{The limit $T \to \infty$.}
On a finite set the scores are bounded, $\lvert z \rvert \le z_{\max}$. A second-order expansion of the exponential gives
\begin{align}
Z_T &= \E_U\Big[1 + \frac{z}{T} + \rho_T(z)\Big],
  \qquad \lvert \rho_T(z) \rvert \le \frac{z_{\max}^2}{2T^2}\, e^{z_{\max}/T}
  && \text{(Taylor's theorem for $e^{z/T}$)} \\
&= 1 + O(T^{-2})
  && \text{($\E_U[z] = 0$)}
\end{align}
so $\log Z_T \to 0$, while $\E_U[z \mid G] / T \to 0$ and $\max_G z / T \to 0$. Both bounds of \cref{prop:temper} therefore tend to 1, and the tempered proposal tends to the uniform one.

\paragraph{Gain and price.}
The lower bound exceeds 1 iff $\E_U[z \mid G] / T > \log Z_T$, that is $\mu_G > T \log Z_T$, which is the gain condition of \cref{sec:rules}.
If $\mu_G \le 0$, the lower bound equals $\exp(-\lvert \mu_G \rvert / T - \log Z_T)$, so acceptance is at least this factor times that of uniform proposals.

\paragraph{Swaps.}
Let a swap $\omega = (a, d)$ draw its add $a$ and its drop $d$ independently, from $Q_+ \propto U_+\, e^{z_+/T_+}$ and $Q_- \propto U_-\, e^{z_-/T_-}$, and let $U = U_+ \times U_-$ be the uniform swap. Then
\begin{align}
w(\omega) = \frac{Q(\omega)}{U(\omega)} &= \frac{Q_+(a)}{U_+(a)} \cdot \frac{Q_-(d)}{U_-(d)}
  && \text{(Independence of the two draws)} \\
&= \exp\Big(\frac{z_+(a)}{T_+} + \frac{z_-(d)}{T_-} - \log Z_{T_+} - \log Z_{T_-}\Big)
  && \text{(Definition of each side's $w$)}
\end{align}
The identity and both bounds above used only $Q = U w$ with $w = \exp(\cdot)$, so they hold for swaps with the exponent replaced by this sum: the exponents of the two sides add. \qed

\subsection{Proof of \cref{prop:fill} (best-response fill)}

Let $h(v) = m\big(x' + e_i(v)\big) - m(x')$, the margin change of adding pixel $i$ at value $v$.
A uniformly random corner $u$ of $B_i$ takes, in each channel independently, each endpoint $-x_{ic}$ and $1 - x_{ic}$ with probability $1/2$.
The best-response corner $v^\ast$ minimizes $\ip{g_i}{v}$ over $B_i$ (\cref{lem:phi}), so in each channel
\begin{align}
g_{ic}\, v^\ast_c &= \min\big\{-g_{ic}\, x_{ic},\; g_{ic}(1 - x_{ic})\big\}
  && \text{(Channel-wise minimizer)} \\
&= g_{ic}\Big(\frac{1}{2} - x_{ic}\Big) - \frac{\lvert g_{ic} \rvert}{2}
  && \text{(Minimum of two numbers $=$ their mean $-$ half their distance $\lvert g_{ic} \rvert$)}
  \label{eq:minpair}
\end{align}
Then
\begin{align}
\E_u[h(u)] - h(v^\ast)
&\ge \E_u\Big[\ip{g_i}{u} - \frac{\lambda_i C}{2}\Big] - \Big(\ip{g_i}{v^\ast} + \frac{\lambda_i C}{2}\Big)
  && \text{(\Cref{lem:reliability} at $u$ and at $v^\ast$)} \\
&= \E_u\ip{g_i}{u} - \ip{g_i}{v^\ast} - \lambda_i C
  && \text{(Linearity of expectation)} \\
&= \sum_{c=1}^{C} \Big(\E_u[g_{ic}\, u_c] - g_{ic}\, v^\ast_c\Big) - \lambda_i C
  && \text{(Inner product as a sum over channels)} \\
&= \sum_{c=1}^{C} \Big(g_{ic}\Big(\frac{1}{2} - x_{ic}\Big) - g_{ic}\, v^\ast_c\Big) - \lambda_i C
  && \text{($\E_u[u_c] = \tfrac{1}{2}(-x_{ic}) + \tfrac{1}{2}(1 - x_{ic})$)} \\
&= \sum_{c=1}^{C} \frac{\lvert g_{ic} \rvert}{2} - \lambda_i C
  && \text{(\Cref{eq:minpair})} \\
&= \frac{\norm{g_i}_1}{2} - \lambda_i C
  && \text{(Definition of the $\ell_1$ norm)}
\end{align}
\qed

\end{document}